%% file: neurips_2024.tex
\documentclass{article}

\usepackage[preprint]{neurips_2026}

\usepackage{enumitem}
\usepackage[utf8]{inputenc} 
\usepackage[T1]{fontenc}    
\usepackage{hyperref}       
\usepackage{url}            
\usepackage{booktabs}       
\usepackage{amsfonts}       
\usepackage{nicefrac}       
\usepackage{microtype}      
\usepackage{graphicx}      
\usepackage{amsmath}
\usepackage{xcolor}
\usepackage{listings}
\usepackage{longtable}
\usepackage{wrapfig}
\usepackage{multirow}
\usepackage[table]{xcolor}
\usepackage{graphicx}
\usepackage{subcaption}
\usepackage{pifont}
\newcommand{\cmark}{\ding{51}}
\newcommand{\xmark}{\ding{55}}

\definecolor{headergray}{gray}{0.92}
\definecolor{bestred}{RGB}{255, 220, 220}
\definecolor{secondgreen}{RGB}{220, 245, 220}

\title{MolViBench: Evaluating LLMs on Molecular Vibe Coding}

\author{%
  Jiatong Li$^{1,2}$ \quad Yuxuan Ren$^2$ \quad Weida Wang$^3$ \quad Changmeng Zheng$^1$ \\
  \textbf{Xiao-yong Wei}$^{1\dagger}$ \quad \textbf{Qing Li}$^1$ \quad \textbf{Yatao Bian}$^2$ \\[0.8ex]
  $^1$The Hong Kong Polytechnic University \quad
  $^2$National University of Singapore \quad
  $^3$Fudan University \\
  $^\dagger$Corresponding author: \texttt{x1wei@polyu.edu.hk} \\
  \url{https://github.com/phenixace/MolViBench-open}
}

\begin{document}

\maketitle

\vskip-0.2in
\input{sections/0_abstract}

\input{sections/1_introduction}
\input{sections/2_relatedwork}
\input{sections/3_benchdesign}
\input{sections/4_expdesign}
\input{sections/5_findings}
\input{sections/6_conclusion}
\bibliography{reference}
\bibliographystyle{rusnat}


\input{sections/appendix}

\end{document}

%% file: sections/0_abstract.tex
\begin{abstract}
\emph{Molecular Vibe Coding}, a paradigm where chemists interact with LLMs to generate executable programs for molecular tasks, has emerged as a flexible alternative to chemical agents with predefined tools, enabling chemists to express arbitrarily complex, customized workflows. Unlike general coding tasks, molecular coding imposes a distinctive challenge that LLMs should jointly equip programming, molecular understanding, and domain-specific reasoning capabilities.
However, existing benchmarks remain disconnected. General code generation benchmarks such as HumanEval and SWE-bench require no chemistry knowledge, while chemistry-focused benchmarks such as S$^2$-Bench and ChemCoTBench evaluate knowledge recall or property prediction rather than executable code generation.
To bridge this gap, we introduce \textbf{MolViBench}, the first benchmark tailored for \emph{Molecular Vibe Coding}. \textbf{MolViBench} comprises 358 curated tasks across five cognitive levels, ranging from single-API recall to end-to-end virtual screening pipeline design, spanning 12 real-world drug discovery workflows.
To rigorously assess generated code, we also propose a multi-layered evaluation framework that combines type-aware output comparison and AST-based API-semantic fallback analysis, which jointly measures executability and chemical correctness. 
We systematically evaluate 9 frontier coding LLMs and compare three real-world \emph{Molecular Vibe Coding} paradigms,
providing a practical and fine-grained testbed for diagnosing LLMs' coding capabilities in AI-accelerated molecular discovery.
\end{abstract}

%% file: sections/1_introduction.tex
\section{Introduction}


\begin{wrapfigure}{r}{0.45\textwidth} 
    \vskip -0.2in
    \centering
    \includegraphics[width=0.43\textwidth]{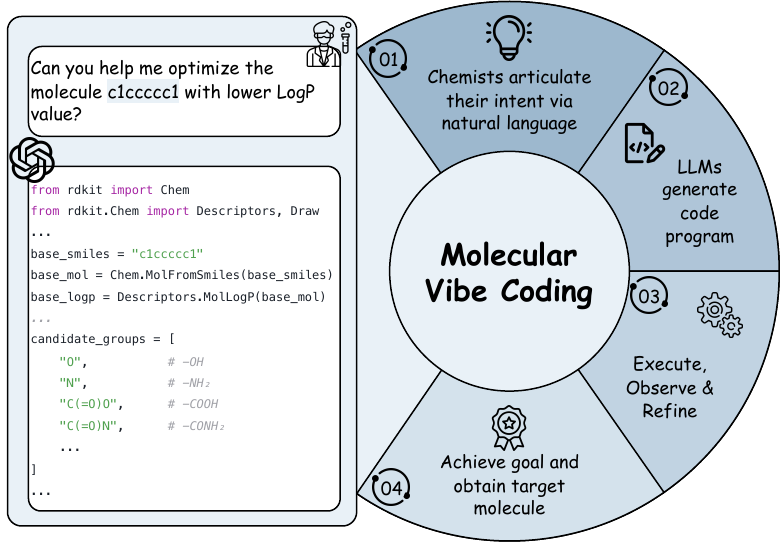} 
    \caption{The illustration of \emph{Molecular Vibe Coding}.}
    \label{fig:intro}
    \vskip -0.1in
\end{wrapfigure}

Large Language Models (LLMs) have demonstrated remarkable potential in scientific research \citep{li2024empowering,wang2025cmphysbench}, especially serving as intelligent assistants that provide insights and recommendations for scientists \citep{li2024speak,solovev2025chemcoscientist}. 
Although chemical agents \citep{m2024augmenting,tang2025chemagent} show promise, they normally adopt fixed function calls with limited extension to flexible operations and emerging requirements.
Recently, as shown in Figure \ref{fig:intro}, a new interaction paradigm, which we name as \emph{Molecular Vibe Coding}, has emerged.
Rather than relying on fixed pipelines or predictive outputs, researchers articulate their intent through natural language, and LLMs synthesize executable programs that perform end-to-end molecular operations under chemical constraints. This paradigm is particularly prevalent in molecular discovery, where chemists and computational biologists could prompt LLMs to compute molecular descriptors \citep{randic1991generalized}, filter compounds by Lipinski's Rule of Five \citep{ivanovic2020lipinski}, or build a virtual screening pipeline \citep{noor2024deep}, expecting functional programs and chemically correct results in return.
This code-centric paradigm unlocks compositional power beyond predefined function calls, enabling chemists to accomplish arbitrarily complex, dynamic workflows through the interaction with LLMs.



Despite the rapid adoption of this paradigm in drug discovery and cheminformatics research, existing benchmarks seldom directly evaluate it. As shown in Table \ref{tab:benchmark_comparison}, current evaluation resources generally fall into two disconnected categories. 
On one hand, general code generation benchmarks such as HumanEval~\citep{chen2021evaluating}, MBPP~\citep{austin2021program}, and SWE-bench~\citep{jimenezswe} measure programming competence through domain-agnostic function synthesis or repository-level bug fixing, but their questions require no chemistry knowledge and impose no chemical correctness constraints. 
On the other hand, chemistry-focused benchmarks such as S$^2$-Bench~\citep{li2024speak} and ChemCoTBench~\citep{haobeyond} focus on knowledge recall or predictive modeling performance, but do not test whether an LLM can produce code that implements molecular computations.
Although works like Biocoder \citep{tang2023biocoder} and ScienceAgentBench \citep{chenscienceagentbench} propose benchmarks on bioinformatics and scientific tasks, they primarily focus on basic tasks rather than real-world analytical pipelines.
This leaves a critical blind spot that the intersection of programming ability, domain-specific chemical reasoning, and output-level chemical correctness remains unevaluated.


\begin{figure}[t]
    \centering
    \vskip -0.25in
    \includegraphics[width=1.0\linewidth]{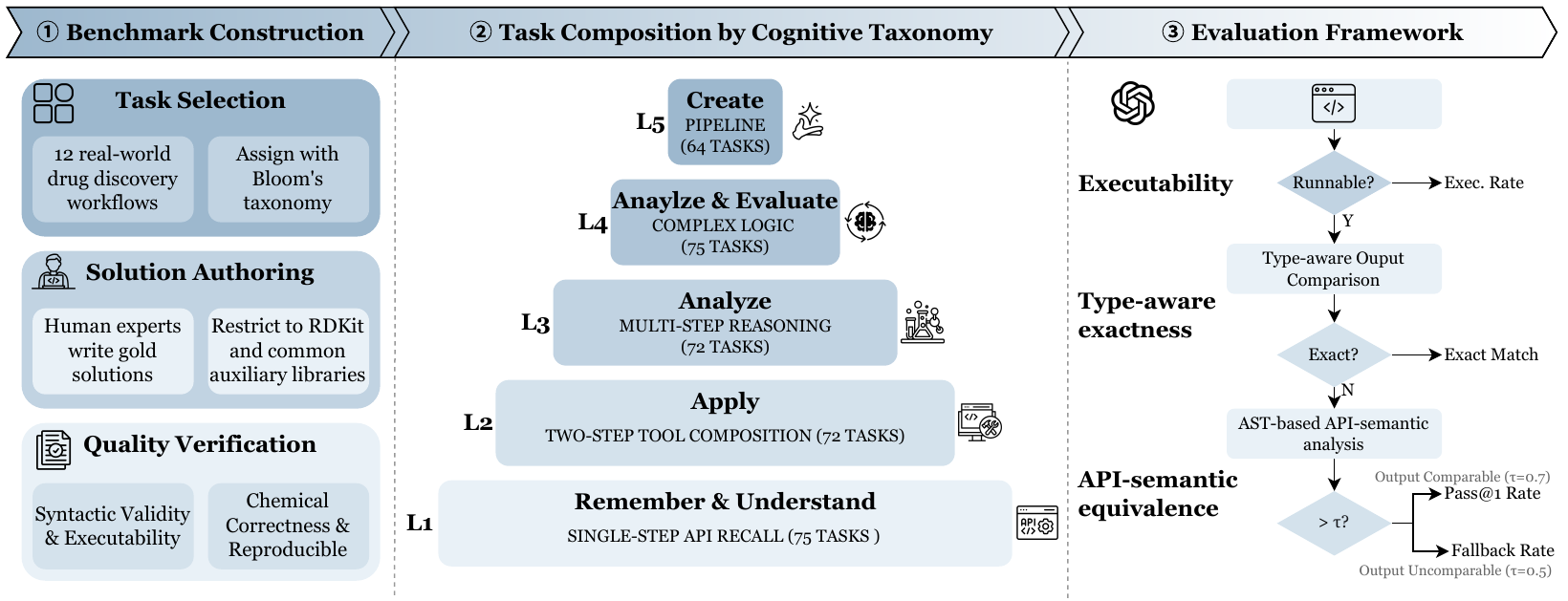}
    \caption{The Overview of MolViBench.}
    \vskip -0.25in
    \label{fig:pipe}
\end{figure}

This gap matters in practice. When an LLM generates code for molecular tasks, the code may execute without errors yet produce chemically invalid results.
For example, a model might correctly call the functions of RDKit \citep{landrum2013rdkit}, but silently mishandle stereochemistry, apply an incorrect reaction template, or yield a molecule that fails to meet the given constraints. 
Such errors are particularly dangerous in scientific workflows because they are difficult to detect without domain expertise, and they can propagate silently into downstream decisions such as compound selection or lead optimization.

Therefore, we introduce \textbf{MolViBench}, the first benchmark tailored for \emph{Molecular Vibe Coding} with 358 real-world molecular coding questions, aiming to test the joint capability among programming, molecular understanding, and domain-specific reasoning, especially for real-world molecule discovery workflows. MolViBench is guided by three design principles:

\noindent\textbf{Progressive Cognitive Levels.} Tasks are stratified into five cognitive levels following Bloom's taxonomy~\citep{krathwohl2002revision}, spanning from basic API recall to end-to-end pipeline design, which mirrors the full cognitive spectrum a chemist exercises during molecular vibe coding. This stratification enables fine-grained diagnosis of capability boundaries, examining whether models falter at memorizing API functions, multi-step chemical reasoning, or orchestrating system-level workflows.

\noindent\textbf{Structured Evaluability.} We enforce a constrained library setting centered on RDKit to eliminate cross-toolkit inconsistencies, and provide reference solutions for all 358 tasks under a unified function interface. Specifically, output comparison is first conducted, extracting and aligning core content across diverse output types, accommodating equivalent representations. For cases where output comparison remains inconclusive, such as stochastic outputs or structurally incomparable return types, we further apply AST-based API-semantic analysis to verify whether the generated code invokes the correct domain functions, serving as a complementary check.

\noindent\textbf{Workflow Realism.} Our questions span 12 categories of real-world drug discovery operations, including molecular characterization, ADMET screening, virtual screening, combinatorial chemistry, lead optimization, reaction simulation, clustering, QSAR modeling, and more. This breadth ensures that benchmark outcomes reflect practical utility rather than artificial difficulty.



We systematically evaluate 9 frontier LLMs and compare three inference paradigms: Direct Generation, Incremental Repair, and Agent Collaboration. Our experiments yield several notable findings. First, Claude Opus 4.6 Think with Incremental Repair achieves the strongest overall performance (Pass@1 = 39.7\%). Second, performance consistently degrades with increasing cognitive level across all models, confirming that higher-order tasks involving multi-step reasoning and workflow orchestration still remain a significant challenge for current coding LLMs. Third, all models fail to surpass 10\% Pass@1 on Level 5 tasks, exposing a fundamental capability gap in synthesizing real-world, end-to-end molecular pipelines.

%% file: sections/2_relatedwork.tex
\section{Related Work}
\subsection{Code Generation Benchmarks}

The evaluation of LLM code generation has progressed through several stages. HumanEval~\citep{chen2021evaluating} and MBPP~\citep{austin2021program} established function-level synthesis benchmarks using unit-test-based pass@$k$ evaluation. SWE-bench~\citep{jimenezswe} extended the scope to repository-level software engineering tasks such as bug fixing and feature implementation. More recent efforts include LiveCodeBench~\citep{jainlivecodebench}, which continuously sources problems from competitive programming platforms to mitigate data contamination, and BigCodeBench~\citep{zhuobigcodebench}, which evaluates practical programming with diverse library usage. DS-1000~\citep{lai2023ds} targets data science code generation with execution-based evaluation.
ScienceAgentBench \citep{chenscienceagentbench} evaluates scientific code generation by requiring executable programs for data-driven tasks.
While these benchmarks have driven significant progress in general-purpose code generation, they do not require domain-specific scientific knowledge. A model can achieve high scores without understanding molecular representations, chemical reactions, or pharmacological constraints. This makes them insufficient for evaluating LLM performance in \emph{Molecular Vibe Coding} scenarios, where correctness depends on both algorithmic logic and domain-specific chemical semantics.

\subsection{Chemistry and Molecular Benchmarks}

Chemistry-focused evaluation resources offer a valuable perspective, while they do not evaluate the coding capability of LLMs directly. For example, MoleculeNet~\citep{wu2018moleculenet} provides a collection of molecular property prediction tasks for benchmarking machine learning models, evaluating predictive accuracy rather than code generation capability. Similarly, Mol-Instructions~\citep{fangmol} and ChemLLMBench~\citep{guo2023can} assess LLM chemical knowledge through instruction-following formats spanning molecular properties, reactions, and molecule design. S$^2$-Bench~\citep{li2024speak} evaluates LLMs on text-based open molecule generation, including editing, optimization, and customized generation. ChemCoTBench~\citep{haobeyond} formalizes chemical reasoning as modular operations to evaluate step-by-step molecular optimization and reaction prediction.
However, many molecular properties, such as exact LogP, QED, or topological polar surface area, are straightforward to compute via established RDKit routines, yet are notoriously difficult and inefficient for LLMs to predict accurately from molecular textual representations.

\input{tables/bench_comparison}




%% file: tables/bench_comparison.tex
\begin{table}[htbp]
\centering
\caption{Comparison of MolViBench with existing code generation and chemistry benchmarks.}
\label{tab:benchmark_comparison}
\resizebox{\textwidth}{!}{%
\begin{tabular}{@{}lcccc@{}}
\toprule
\rowcolor{headergray}
\textbf{Benchmark} & \textbf{Domain} & \textbf{Code Gen.} & \textbf{Domain Know.} & \textbf{Exec.-Based Eval.} \\
\midrule
HumanEval~\citep{chen2021evaluating}          & General        & \cmark & \xmark & \cmark \\
MBPP~\citep{austin2021program}                & General        & \cmark & \xmark & \cmark \\
SWE-bench~\citep{jimenezswe}                  & Software Eng.  & \cmark & \xmark & \cmark \\
LiveCodeBench~\citep{jainlivecodebench}       & General        & \cmark & \xmark & \cmark \\
BigCodeBench~\citep{zhuobigcodebench}         & General        & \cmark & \xmark & \cmark \\
DS-1000~\citep{lai2023ds}                     & Data Science   & \cmark & \xmark & \cmark \\
ScienceAgentBench~\citep{chenscienceagentbench} & Science      & \cmark & \cmark & \cmark \\
\midrule
MoleculeNet~\citep{wu2018moleculenet}         & Chemistry      & \xmark & \cmark & \xmark \\
Mol-Instructions~\citep{fangmol}              & Chemistry      & \xmark & \cmark & \xmark \\
S$^2$-Bench~\citep{li2024speak}               & Chemistry      & \xmark & \cmark & \xmark \\
ChemCoTBench~\citep{haobeyond}                & Chemistry      & \xmark & \cmark & \xmark \\
ChemLLMBench~\citep{guo2023can} & Chemistry & \xmark & \cmark & \xmark \\
\midrule
\textbf{MolViBench (Ours)}                    & Chemistry      & \cmark & \cmark & \cmark \\
\bottomrule
\end{tabular}%
}
\end{table}

%% file: sections/3_benchdesign.tex
\section{MolViBench}
\label{sec:benchmark}


MolViBench includes 358 real-world molecular coding tasks that simulates the \emph{Molecular Vibe Coding} workflow. In this paradigm, a chemist provides a natural-language instruction, and the LLM must generate code that is both executable and chemically accurate to accomplish the molecular task. Each task requires the model to produce a self-contained Python function that processes molecular inputs and returns a chemically meaningful result. In this work, we restrict the generated code to RDKit as the primary cheminformatics library, with NumPy, pandas, scikit-learn, matplotlib, and selfies permitted as auxiliary dependencies.

\subsection{Task Composition}
\label{sec:bloom}

As demonstrated in Figure \ref{fig:pipe}, MolViBench organizes all 358 tasks into five cognitive levels with increasing difficulty following Bloom's taxonomy~\citep{krathwohl2002revision}, aiming to test the joint capability among programming, molecular understanding, and domain-specific reasoning.

\noindent\textbf{Level 1: Remember \& Understand (75 tasks).} L1 tasks require basic API recall and single-step operations. For example, molecular descriptor computation (molecular weight, LogP, TPSA), format conversion (SMILES to InChI, SMILES to MOL block), substructure queries, and simple property checks. These tasks test whether the model is familiar with common RDKit function signatures and can apply them correctly.

\noindent\textbf{Level 2: Apply (72 tasks).} L2 tasks involve composing two or more API calls to perform intermediate transformations, including fingerprint generation followed by similarity computation, BRICS decomposition, maximum common substructure (MCS) extraction, Butina clustering, and conformer generation with energy minimization. Success in level 2 tasks requires understanding of molecular knowledge and how RDKit objects flow between functions.

\noindent\textbf{Level 3: Analyze (72 tasks).} Tasks in level 3 demand multi-step chemical reasoning that goes beyond API composition, such as reaction simulation (Suzuki coupling, Click chemistry, Buchwald--Hartwig amination), retrosynthetic analysis, ADMET-related computations (CYP450 metabolism prediction, hERG liability assessment), and structural alert detection (PAINS, Brenk filters). LLMs must combine programming logic with domain-specific chemical knowledge.

\noindent\textbf{Level 4: Analyze \& Evaluate (75 tasks).} Tasks in level 4 require conditional branching, iterative optimization, and complex control flow under chemical constraints. We identify seven distinct reasoning patterns in this level, including backtracking search, error recovery, and multi-molecule coordination. These tasks probe whether models can generate robust code that handles edge cases and adapts to intermediate results.

\noindent\textbf{Level 5: Create (64 tasks).} L5 tasks require designing and implementing end-to-end computational pipelines. For instance, virtual screening workflows (library enumeration $\to$ filtering $\to$ scoring $\to$ ranking), genetic algorithm-based molecular optimization, QSAR model construction with scaffold-split cross-validation, and multi-objective Pareto optimization. These tasks test system-level integration and the ability to orchestrate multiple computational stages into a coherent pipeline.

Unlike traditional difficulty scales based on code length or test-case count, our hierarchy is \emph{cognition-driven}, which means that an L5 task demands high-level cognitive capabilities like reasoning, planning, and molecular understanding rather than naive recall and application of RDKit functions.

\subsection{Benchmark Construction}
\label{sec:construction}

The construction of MolViBench follows a three-stage pipeline: \emph{task design}, \emph{solution authoring}, and \emph{quality assurance}.

\noindent\textbf{Task selection.}
We first survey real-world cheminformatics workflows in drug discovery literature and identify 12 representative operation categories, as listed in Table~\ref{tab:workflow_coverage}. For each category, domain experts with backgrounds in computational chemistry and cheminformatics manually compose natural-language instructions that reflect authentic research needs. 
Each task is then assigned to a Bloom's taxonomy level based on the cognitive operations it requires. To ensure balanced coverage, we iteratively adjust the task pool until every level-category combination that is semantically meaningful contains at least one task.

\noindent\textbf{Solution authoring.}
For every task, we author a gold-standard Python reference solution that follows a unified function interface: a single entry point \texttt{level\_function()} that accepts molecular inputs (typically SMILES strings or molecule lists) and returns a chemically meaningful result. All solutions are restricted to RDKit as the primary cheminformatics library, with NumPy, pandas, scikit-learn, matplotlib, and selfies permitted as auxiliary dependencies. This single-library constraint eliminates cross-toolkit inconsistencies. For example, RDKit and Open Babel may produce different canonical SMILES for the same molecule, which would undermine deterministic evaluation. 

\noindent\textbf{Quality verification.}
Each task-solution pair undergoes a two-round verification process. In the first round, all solutions are executed programmatically to confirm syntactic validity and adherence to the unified function interface, including correct input/output signatures. In the second round, an independent reviewer executes each reference solution against a standardized test molecule set, comprising 20 single molecules, 10 molecule pairs, and a 50-molecule library, and verifies that the outputs are chemically correct and reproducible across inputs. Tasks that fail either round are revised or removed. After quality verification, the final benchmark comprises 358 tasks with verified reference solutions.

\subsection{Statistics}
\label{sec:statistics}

\input{tables/satistic}


Table~\ref{tab:level_stats} summarizes the distribution of tasks across cognitive levels and output types.
Several patterns emerge from the distribution. Lower levels (L1 \& L2) are dominated by simple output types (\texttt{int}, \texttt{float}, \texttt{bool}, \texttt{str}), reflecting single-value computations such as molecular weight or substructure matching. As cognitive complexity increases, structured output types (\texttt{list}, \texttt{dict}) become overwhelmingly prevalent. Level~4 tasks exclusively return \texttt{dict}, encoding multi-field results from complex conditional and iterative workflows, while Level~5 tasks return either \texttt{list} or \texttt{dict} to capture pipeline outputs such as ranked candidate lists or comprehensive screening reports. 

%% file: tables/satistic.tex
\begin{table}[t]
\centering
\vskip -0.25in
\caption{Task distribution across five Bloom's taxonomy levels and output types. ``Other'' includes \texttt{set}, \texttt{Image}, and \texttt{DataFrame} returns.}
\label{tab:level_stats}
\begin{tabular}{@{}ccccccccccc@{}}
\toprule
\rowcolor{headergray}
\textbf{Level} & \textbf{Tasks} & \texttt{int} & \texttt{float} & \texttt{bool} & \texttt{str} & \texttt{list} & \texttt{dict} & \texttt{tuple} & Other \\
\midrule
L1    & 75 & 19 & 12 & 21 & 8 & 5  & 6 & 1 & 3 \\
L2      & 72 & 1  & 5  & 3  & 29 & 24 & 7 & 1 & 2 \\
L3    & 72 & 0  & 0  & 6  & 7  & 34 & 25 & 0 & 0 \\
L4   & 75 & 0  & 0  & 0  & 0  & 0  & 75 & 0 & 0 \\
L5    & 64 & 0  & 0  & 0  & 0  & 42 & 22 & 0 & 0 \\
\midrule
\rowcolor{headergray}
\textbf{Total}  & \textbf{358} & 20 & 17 & 30 & 44 & 105 & 135 & 2 & 5 \\
\bottomrule
\end{tabular}

\end{table}

%% file: sections/4_expdesign.tex
\section{Experiments}
\label{sec:experiments}

In this section, we describe the experimental setup, including the benchmarked models, the inference paradigms, and the evaluation framework, as well as the metrics used to assess both executability and chemical correctness.

\subsection{Benchmarked Models}

We evaluate 9 frontier LLMs spanning six providers. Table~\ref{tab:models} lists all models along with the inference paradigms under which each is tested. The selection covers the current state of the art in code generation, including general-purpose models (Gemini 3 Pro \citep{gemini3}, DeepSeek-V3.2 Chat \citep{liu2025deepseek}), code-specialized models (Claude Opus 4.6 \citep{anthropic_claude_2026}, GPT-5.2/5.3 Codex \citep{openai_codex52,openai_codex53}), and reasoning models (Claude Opus 4.6 Thinking \citep{anthropic_claude_2026}, DeepSeek-V3.2 Reasoner \citep{liu2025deepseek}). 

\input{tables/models}

For each problem of MolViBench, we provide the natural-language prompt under a standardized system instruction that specifies the coding constraints: the generated function must be named \texttt{level\_function}, use RDKit as the primary library, wrap the body in \texttt{try/except}, and return \texttt{None} for invalid inputs. We adopt a basic prompt strategy throughout that only the task description is given, with no function signatures, input-output examples, or few-shot demonstrations, representing the most realistic and challenging evaluation setting.

\subsection{Inference Paradigms}
\label{sec:paradigms}

Beyond single-turn direct generation, we investigate whether increasingly sophisticated interaction paradigms can enhance the performance of \emph{Molecular Vibe Coding}. Here, we explain the difference among three distinct inference paradigms:

\noindent\textbf{Direct generation.} The model receives the task prompt and produces code in a single forward pass with no execution feedback. This paradigm serves as the baseline and reflects the simplest deployment scenario.

\noindent\textbf{Incremental Repair.} After the model generates an initial solution, the code is executed locally. If execution fails due to syntax errors, runtime exceptions, or incorrect output types, the error traceback is appended to the conversation history and the model is prompted to revise its code.

\noindent\textbf{Agent Collaboration.} Two LLM instances collaborate in a Coder-Tester protocol. One agent (Coder) generates the solution, while the other agent (Tester) independently designs a test plan by selecting test molecules, specifying expected output types and reasonable value ranges, and defining failure criteria.  The Tester executes the Coder’s output, evaluates chemical plausibility, and returns structured feedback (i.e., PASS/FAIL with explanation). Upon failure, this feedback is relayed to the Coder for revision. This cycle repeats until the code executes successfully or the maximum number of interaction rounds is reached.

\subsection{Evaluation Framework}
\label{sec:eval_framework}

Evaluating \emph{Molecular Vibe Coding} is fundamentally challenging due to (i)~\emph{compositional complexity}: multi-step chemical reasoning with heterogeneous output, and (ii)~\emph{solution multiplicity}: multiple valid outputs and algorithmic pathways. These properties render exact output matching insufficient as a standalone criterion.

We formalize our evaluation framework as a three-stage pipeline. For each task $q \in \mathcal{Q}$, we are given a reference implementation $f_q^\star$ and a task-specific semantic validator $\mathcal{L}_q$. Given a model-generated program $\hat{f}_q$, we define the following stages:

\noindent\textbf{Stage 1: Executability.}
We first assess whether $\hat{f}_q$ constitutes a valid program. This includes (i)~syntactic validity via \texttt{compile()}, (ii)~presence of the required entry point (\texttt{level\_function}) via AST inspection, and (iii)~successful execution on a set of inputs $\{x_i\}_{i=1}^5$ under a 30\,s timeout. We define
\[
\mathrm{Exec}(q) = \mathbb{1}\big[\forall i,\; \hat{f}_q(x_i)\ \text{executes without error}\big],
\]
and report the \textbf{Executable rate} as $\mathbb{E}_{q}[\mathrm{Exec}(q)]$.

\noindent\textbf{Stage 2: Type-aware exactness.}
For executable programs, we evaluate output exactness under a type-dispatched equivalence relation $\sim_{\text{type}}$. This comparator accounts for heterogeneous scientific outputs: exact matching for discrete types, tolerance-based comparison for floating-point values, canonical SMILES equivalence for molecular strings, and recursive matching for structured objects. For non-deterministic tasks, we replace value matching with structural constraints (type, shape, key set).

A task is considered an exact match only if all test inputs satisfy the equivalence:
\[
\mathrm{EM}(q) = \mathbb{1}\big[\forall i \in \mathcal{I}_q,\; \hat{f}_q(x_i) \sim_{\text{type}} f_q^\star(x_i)\big].
\]

We report \textbf{EM} as $\mathbb{E}_q[\mathrm{EM}(q)]$.

\noindent\textbf{Stage 3: API-semantic equivalence.}
Exact matching may fail when the outputs are uncomparable. Meanwhile, for high-level questions, there can be multiple valid solutions. To capture the two situations, we define a relaxed notion of equivalence based on API semantics. Let $\mathcal{A}(f)$ denote the set of RDKit functional categories extracted from program $f$ via AST parsing. We exclude non-discriminative primitives and merge functionally equivalent APIs.

We define API coverage as
\[
\mathrm{Cov}(q) = \frac{|\mathcal{A}(\hat{f}_q) \cap \mathcal{A}(f_q^\star)|}{|\mathcal{A}(f_q^\star)|}.
\]
A program satisfies API-semantic correctness if $\mathrm{Cov}(q) \geq \tau$ and $\mathrm{Exec}(q)=1$.

We apply this check under two regimes that differ in \emph{when} the fallback is triggered and \emph{how strict} the threshold is:

\begin{itemize}[nosep,leftmargin=1.5em]
    \item \textbf{Pass@1 Rate} activates the API check only when exact comparison is \emph{structurally infeasible}(i.e., the reference and prediction return incomparable types), such as an image object vs.\ an SVG string, a 3D conformer vs.\ a coordinate matrix. In such cases we require $\mathrm{Cov}(q) \geq 0.5$:
    \[
    \mathrm{Pass}(q) = \mathrm{Exact}(q) \;\lor\; \bigl[\,\lnot\mathrm{Comp}(q) \;\land\; \mathrm{Cov}(q) \geq 0.5\,\bigr],
    \]
    where $\mathrm{Comp}(q)$ indicates that the output types are comparable. We report $\mathbb{E}_q[\mathrm{Pass}(q)]$ as \textbf{Pass@1}, which serves as our \textbf{main pass metric} and extends exact matching to handle type-incomparable outputs while keeping a moderate overlap bar.

    \item \textbf{Fallback Pass Rate} (\textbf{FB}) activates the API check for \emph{all} executable predictions that fail exact matching, regardless of type comparability. To compensate for the broader trigger, we raise the threshold to $\mathrm{Cov}(q) \geq 0.7$:
    \[
    \mathrm{Fallback}(q) = \mathrm{Exact}(q) \;\lor\; \bigl[\,\mathrm{Exec}(q) \;\land\; \mathrm{Cov}(q) \geq 0.7\,\bigr].
    \]
    We report $\mathbb{E}_q[\mathrm{Fallback}(q)]$ as \textbf{Fallback Pass Rate} to capture solutions that invoke correct RDKit workflows but produce different outputs, where multiple valid solution paths exist.
\end{itemize}





%% file: tables/models.tex
\begin{table}[t]
\centering
\caption{LLMs evaluated in MolViBench. All models are accessed via their official APIs. ``IR'' denotes the Incremental Repair paradigm, while ``AC'' denotes the Agent Collaboration paradigm.}
\label{tab:models}
\begin{tabular}{@{}llll@{}}
\toprule
\rowcolor{headergray}
\textbf{Model} & \textbf{Provider} & \textbf{Type} & \textbf{Paradigms} \\
\midrule
Claude Opus 4.6 \citep{anthropic_claude_2026} & Anthropic & Code-specialized & Direct, IR, AC \\
Claude Opus 4.6 Thinking \citep{anthropic_claude_2026} & Anthropic & Reasoning & Direct, IR, AC \\
Gemini 3 Pro \citep{gemini3} & Google & Standard & Direct, IR, AC \\
DeepSeek-V3.2 Reasoner \citep{liu2025deepseek} & DeepSeek & Reasoning & Direct, IR, AC \\
DeepSeek-V3.2 Chat \citep{liu2025deepseek} & DeepSeek & Standard & Direct, IR \\
GPT-5.2 Codex \citep{openai_codex52} & OpenAI & Code-specialized & Direct, IR \\
GPT-5.3 Codex \citep{openai_codex53} & OpenAI & Code-specialized & Direct, IR \\
Kimi-K2.5 \citep{team2026kimi} & Moonshot & Standard & Direct \\
MiniMax-M2.5 \citep{minimax25} & MiniMax & Standard & Direct \\
\bottomrule
\end{tabular}

\end{table}

%% file: sections/5_findings.tex
\section{Discussion}
\label{sec:findings}

\subsection{Overall Performance}

\input{tables/main_table}

Table~\ref{tab:main_results} presents the overall performance of different models under various inference paradigms on MolViBench, ranked by Pass@1.

Overall, Claude Opus 4.6 Think with Incremental Repair achieves the strongest overall performance (Pass@1 39.7\%, executable rate 98.9\%, Fallback Pass Rate 72.6\%), and notably, all top-four entries are Claude Opus 4.6 variants, indicating that both intrinsic model capability and post-hoc repair strategy compound to drive performance gains.
Meanwhile, across all models, Pass@1 remains below 39.2\%, and performance degrades monotonically across the five cognitive levels, with every model scoring under 10\% on Level 5 tasks, revealing that end-to-end pipeline synthesis remains a fundamental bottleneck even for frontier models that achieve strong human-evaluation scores. These results underscore the necessity of a dedicated benchmark for \emph{Molecular Vibe Coding}, where general coding proficiency does not transfer to chemically grounded, multi-step workflow generation.

\subsection{Per Level Analysis}

\begin{wrapfigure}{r}{0.75\textwidth}
    \centering
    \includegraphics[width=1.0\linewidth]{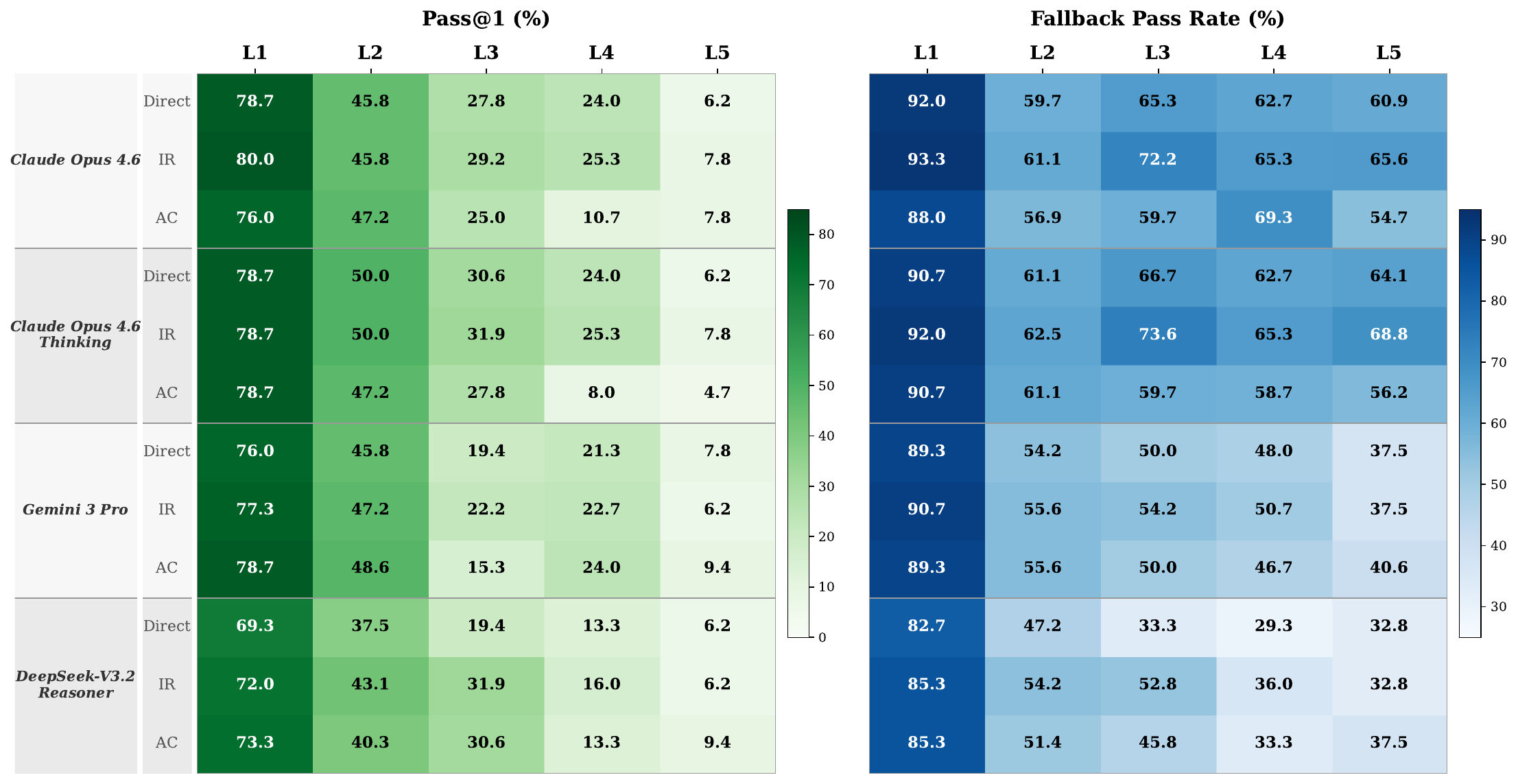}
    \caption{Per-level performance heatmap across inference paradigms and LLMs. Left: Pass@1 (\%); Right: Fallback Pass Rate (\%).}
    \label{fig:heat}
\end{wrapfigure}

Figure~\ref{fig:heat} analyzes model behavior across Bloom’s taxonomy levels to understand how task complexity affects performance and failure modes.


\textbf{Performance degrades with increasing cognitive level.}
Across all models, Pass@1 consistently decreases from L1 to L5, indicating that tasks aligned with higher-order cognitive processes are significantly more challenging. L1 tasks, which primarily involve direct API usage and simple compositions, are handled reliably by all models. In contrast, L3-L5 tasks require multi-step reasoning, conditional logic, and workflow orchestration, where performance drops substantially. This trend validates that MolViBench effectively stratifies tasks along increasing levels of difficulty.

\textbf{High-level tasks expose execution and orchestration failures.}
While Pass@1 decreases at higher levels, Fallback Pass Rate (FB) remains relatively high for several models at L4 and L5. This discrepancy indicates that models are often able to select appropriate APIs and follow semantically valid solution strategies, but fail to produce outputs that align with the reference implementation. In other words, models frequently generate partially correct solutions that fail due to execution issues such as improper sequencing, missing intermediate steps, or inconsistencies in program structure.

\textbf{Model-specific differences emerge at higher levels.}
We also observe variation across models in how performance degrades. Stronger reasoning models (e.g., Claude Opus 4.6 Thinking) maintain relatively higher performance at mid-level tasks, but still experience significant drops at L4-L5. Code-specialized models (e.g., GPT-based Codex variants) perform competitively at intermediate levels but struggle to maintain consistency at higher levels. These differences suggest that while models may excel in either reasoning or code generation individually, combining both capabilities in long-horizon tasks remains challenging.

\subsection{Inference Paradigm Comparison}

\begin{figure}
    \centering
    \includegraphics[width=1.0\linewidth]{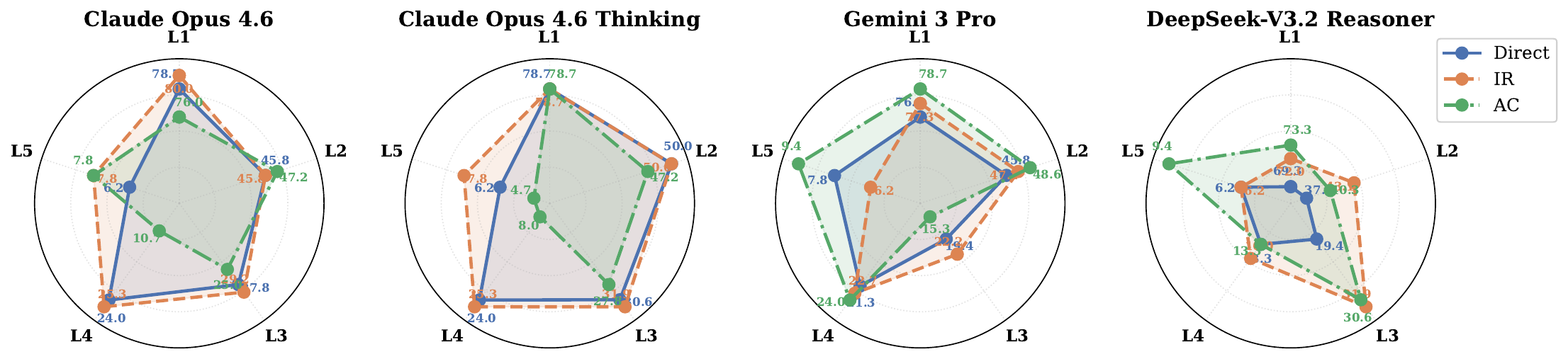}
    \caption{Comparison of Three Inference Paradigms (Direct = Direct Generation; IR = Incremental Repair; AC = Agent Collaboration) on MolViBench.}
    \label{fig:para}
\end{figure}

Figure~\ref{fig:para} compares different inference paradigms to understand how post-hoc refinement strategies interact with model capabilities across difficulty levels.

\textbf{Agent Collaboration is sensitive to model quality.}
Agent Collaboration (AC) achieves comparable or even improved Pass@1 for relatively weaker models (e.g., Gemini 3 Pro and DeepSeek-V3.2 Reasoner), but tends to degrade performance for stronger models such as Claude Opus 4.6. This suggests that the effectiveness of holistic rewriting is highly dependent on the underlying model's capability.
For weaker models, AC can provide useful global restructuring and error correction, leading to performance gains. In contrast, for stronger models that already produce largely correct solutions, the coder-tester paradigm may introduce unnecessary modifications that disrupt correct intermediate reasoning, resulting in performance regression. 

\textbf{Incremental Repair is consistently robust across models and levels.}
In contrast to AC, Incremental Repair (IR) consistently improves performance across models without introducing regressions. IR achieves stable gains at both lower and higher levels, indicating that localized corrections to non-executable or partially incorrect code are more reliable than full-solution regeneration. This robustness holds across different model families, suggesting that IR is less sensitive to the underlying model capability and more aligned with preserving correct intermediate structure.

\textbf{Paradigm effectiveness depends on interaction with task difficulty.}
The relative performance of paradigms also varies with Bloom levels. AC tends to perform better on simpler tasks but degrades at higher levels, where preserving partial correctness becomes critical. IR, on the other hand, maintains consistent gains across levels, and is particularly effective in settings where execution errors are prevalent. These observations indicate that inference paradigms interact with both model capability and task complexity in non-trivial ways.


%% file: tables/main_table.tex
\begin{table}[t]
\centering
\caption{Full results (\%) on MolViBench across all inference paradigms. Models ranked by Pass@1 rate. Exec.\,=\,executable rate; EM\,=\,Exact Match; P@1\,=\,Pass@1; FB\,=\,Fallback Pass Rate. L1--L5 report per-level Pass@1. Best results are highlighted in \colorbox{bestred}{red}, while second-best results are highlighted in \colorbox{secondgreen}{green}.}
\label{tab:main_results}
\setlength{\tabcolsep}{3.5pt}
\begin{tabular}{@{}lccccccccc@{}}
\toprule
\rowcolor{headergray}
\textbf{Model} & \textbf{Exec.} & \textbf{EM} & \textbf{P@1} & \textbf{FB} & \textbf{L1} & \textbf{L2} & \textbf{L3} & \textbf{L4} & \textbf{L5} \\
\midrule
Claude Opus 4.6 Think (IR)
& \cellcolor{bestred}98.9 & \cellcolor{bestred}33.2 & \cellcolor{bestred}39.7 & \cellcolor{bestred}72.6
& \cellcolor{secondgreen}78.7 & \cellcolor{secondgreen}50.0 & \cellcolor{bestred}31.9 & \cellcolor{bestred}25.3 & \cellcolor{secondgreen}7.8 \\

Claude Opus 4.6 Think
& 94.1 & \cellcolor{secondgreen}32.7 & \cellcolor{secondgreen}38.8 & 69.3
& \cellcolor{secondgreen}78.7 & \cellcolor{secondgreen}50.0 & \cellcolor{secondgreen}30.6 & \cellcolor{secondgreen}24.0 & 6.2 \\

Claude Opus 4.6 (IR)
& \cellcolor{secondgreen}98.6 & 32.1 & 38.6 & \cellcolor{secondgreen}71.8
& \cellcolor{bestred}80.0 & 45.8 & 29.2 & \cellcolor{bestred}25.3 & \cellcolor{secondgreen}7.8 \\

Claude Opus 4.6
& 94.1 & 31.0 & 37.4 & 68.4
& \cellcolor{secondgreen}78.7 & 45.8 & 27.8 & \cellcolor{secondgreen}24.0 & 6.2 \\

Gemini 3 Pro (IR)
& \cellcolor{secondgreen}98.6 & 30.4 & 36.0 & 58.4
& 77.3 & 47.2 & 22.2 & 22.7 & 6.2 \\

Gemini 3 Pro (AC)
& 95.5 & 31.0 & 36.0 & 57.0
& \cellcolor{secondgreen}78.7 & 48.6 & 15.3 & \cellcolor{secondgreen}24.0 & \cellcolor{bestred}9.4 \\

Gemini 3 Pro
& 94.1 & 29.3 & 34.9 & 56.4
& 76.0 & 45.8 & 19.4 & 21.3 & \cellcolor{secondgreen}7.8 \\

DeepSeek-V3.2 Reasoner (IR)
& 86.6 & 28.8 & 34.6 & 52.8
& 72.0 & 43.1 & \cellcolor{bestred}31.9 & 16.0 & 6.2 \\

Claude Opus 4.6 Think (AC)
& 95.2 & 28.5 & 34.1 & 65.6
& \cellcolor{secondgreen}78.7 & 47.2 & 27.8 & 8.0 & 4.7 \\

Claude Opus 4.6 (AC)
& 95.0 & 26.5 & 34.1 & 66.2
& 76.0 & 47.2 & 25.0 & 10.7 & \cellcolor{secondgreen}7.8 \\

DeepSeek-V3.2 Reasoner (AC)
& 88.5 & 28.8 & 34.1 & 51.1
& 73.3 & 40.3 & \cellcolor{secondgreen}30.6 & 13.3 & \cellcolor{bestred}9.4 \\

GPT-5.2 Codex
& 89.9 & 27.9 & 33.2 & 51.1
& 76.0 & 44.4 & 23.6 & 9.3 & \cellcolor{bestred}9.4 \\

GPT-5.3 Codex
& 89.7 & 27.9 & 32.7 & 64.8
& 76.0 & \cellcolor{bestred}55.6 & 16.7 & 5.3 & 6.2 \\

MiniMax M2.5
& 84.9 & 24.6 & 30.4 & 50.6
& 64.0 & 40.3 & 26.4 & 14.7 & 3.1 \\

DeepSeek-V3.2 Reasoner
& 74.6 & 24.9 & 29.9 & 45.5
& 69.3 & 37.5 & 19.4 & 13.3 & 6.2 \\

DeepSeek-V3.2 Chat
& 84.6 & 20.9 & 28.2 & 48.9
& 68.0 & 40.3 & 15.3 & 6.7 & \cellcolor{secondgreen}7.8 \\

Kimi K2.5
& 64.2 & 22.4 & 27.9 & 36.9
& 74.7 & 43.1 & 6.9 & 6.7 & 4.7 \\
\bottomrule
\end{tabular}
\vskip -0.2in
\end{table}

%% file: sections/6_conclusion.tex
\section{Conclusion}

We present MolViBench, the first benchmark for evaluating LLM in \emph{Molecular Vibe Coding}. This increasingly prevalent workflow allows researchers to describe molecular computation tasks in natural language and rely on LLMs to generate executable cheminformatics code. MolViBench comprises 358 tasks organized into five progressive cognitive levels spanning 12 drug discovery workflow categories, accompanied by a multi-layered deterministic evaluation framework that jointly measures the programming, molecular understanding, and domain-specific reasoning capabilities to accomplish complex molecule discovery workflows. Our comprehensive benchmarking on 9 frontier LLMs shows that current models, while excellent in single-step API function calls, still face significant challenges in high level molecular tasks that require complex reasoning and multi-step logic.
MolViBench provides systematic diagnosis of LLM capability across the full cognitive spectrum of molecular vibe coding, offering actionable guidance for the development of chemical coding LLMs.

%% file: sections/appendix.tex
\newpage
\appendix

\definecolor{promptbg}{RGB}{245,245,250}
\definecolor{promptframe}{RGB}{180,180,200}
\definecolor{codebg}{RGB}{248,248,248}

\lstdefinestyle{promptstyle}{
    basicstyle=\small\ttfamily,
    backgroundcolor=\color{promptbg},
    frame=single,
    rulecolor=\color{promptframe},
    breaklines=true,
    breakatwhitespace=false,
    columns=fullflexible,
    keepspaces=true,
    aboveskip=6pt,
    belowskip=6pt,
    xleftmargin=4pt,
    xrightmargin=4pt,
}

\lstdefinestyle{pythonstyle}{
    language=Python,
    basicstyle=\scriptsize\ttfamily,
    backgroundcolor=\color{codebg},
    frame=single,
    rulecolor=\color{promptframe},
    breaklines=true,
    breakatwhitespace=false,
    columns=fullflexible,
    keepspaces=true,
    aboveskip=6pt,
    belowskip=6pt,
    xleftmargin=4pt,
    xrightmargin=4pt,
    showstringspaces=false,
    keywordstyle=\color{blue},
    commentstyle=\color{gray},
    stringstyle=\color{red!70!black},
}

\section{Hyperparameter Settings}
\label{app:hyperparams}

Table~\ref{tab:hyperparams} summarizes the hyperparameters used across all inference paradigms.

\begin{table}[h]
\centering
\caption{Hyperparameter settings for each inference paradigm.}
\label{tab:hyperparams}
\small
\begin{tabular}{lccc}
\toprule
\rowcolor{headergray}
\textbf{Parameter} & \textbf{Direct} & \textbf{Incremental Repair} & \textbf{Agent Collaboration} \\
\midrule
Temperature & 0.0 & 0.0 & 0.0 \\
Max tokens & 4096 & 4096 & 4096 \\
API timeout (s) & 120 & 120 & 120 \\
Execution timeout (s) & --- & 30 & 30 \\
Max rounds & --- & 3 & 3 \\
\bottomrule
\end{tabular}
\end{table}

All experiments use greedy decoding (temperature = 0) to ensure reproducibility. Incremental Repair tests generated code against representative molecules from the ZINC250K dataset~\citep{sterling2015zinc} and feeds runtime errors back to the model for up to 3 repair rounds. The Agent Collaboration paradigm uses lower concurrency due to its higher per-question API call count (2--7 calls per question depending on repair rounds).

\section{Sensitivity Analysis of Fallback Pass Rate}

\begin{figure}[h]
    \centering
    \includegraphics[width=0.75\linewidth]{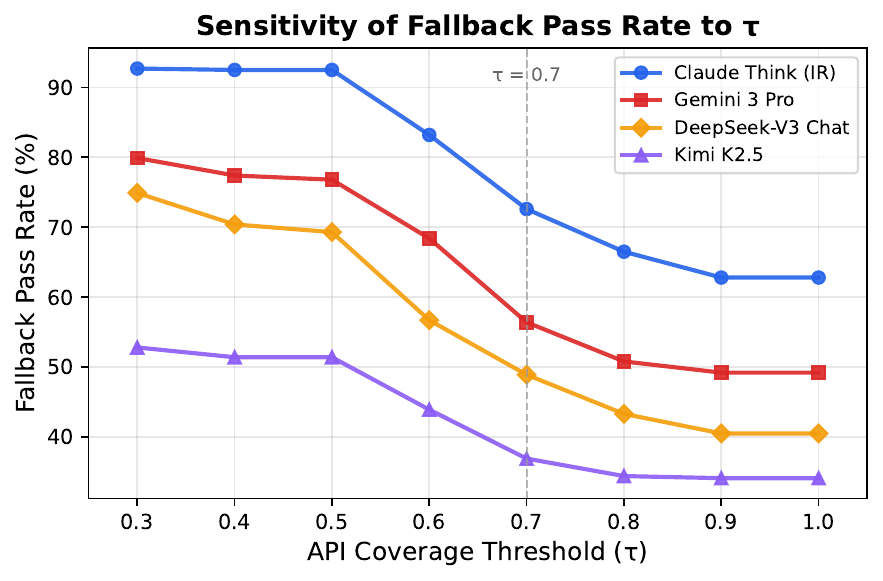}
    \caption{Fallback Pass Rate as a function of the API-coverage threshold $\tau$. The dashed vertical line marks $\tau=0.7$ adopted in the main evaluation. All models exhibit a clear inflection between $\tau=0.6$ and $\tau=0.8$.}
    \label{fig:tau_sensitivity}
\end{figure}

Figure~\ref{fig:tau_sensitivity} and Table~\ref{tab:tau_sensitivity} report how Fallback Pass Rate (FB) varies with the API-coverage threshold $\tau$ across four representative models.

\paragraph{Threshold selection.}
All models exhibit a monotonic decline in FB as $\tau$ increases. The steepest drop occurs between $\tau=0.6$ and $\tau=0.8$, with the largest single-step decrease concentrated at $\tau=0.6\to0.7$ (Claude~Think~IR: $-$10.6\,pp; Gemini~3~Pro: $-$12.0\,pp; DeepSeek-V3: $-$7.8\,pp; Kimi~K2.5: $-$7.0\,pp), marking this as a natural decision boundary. We therefore select $\tau=0.7$, which (i)~lies within the transition region, (ii)~provides meaningful model separation (spread of $\sim$36\,pp), and (iii)~maintains non-trivial selectivity without being overly permissive.

\paragraph{Boundary behaviour.}
At $\tau=0.3$, thresholds are permissive to the point of being uninformative: even the weakest model (Kimi~K2.5) reaches 52.8\%. Conversely, at $\tau \geq 0.9$ the curves flatten, indicating that requiring near-perfect API overlap excludes very few additional predictions and adds little discriminative power.

\paragraph{FB vs.\ EM at strict overlap.}
At $\tau=1.0$, FB remains substantially above Exact Match for all models (e.g., Claude~Think~IR: 62.8\% vs.\ 33.2\% EM). This confirms that a large fraction of EM failures reflect output-format divergence rather than methodological errors. Predictions that use \emph{exactly} the reference API set but produce outputs that do not match the gold string.

\begin{table}[h]
\centering
\caption{Fallback Pass Rate (\%) at varying $\tau$ thresholds. Underlined values correspond to $\tau=0.7$ used in the main paper.}
\label{tab:tau_sensitivity}
\small
\begin{tabular}{@{}lcccccccc@{}}
\toprule
\rowcolor{headergray}
\textbf{Model} & \textbf{0.3} & \textbf{0.4} & \textbf{0.5} & \textbf{0.6} & \textbf{0.7} & \textbf{0.8} & \textbf{0.9} & \textbf{1.0} \\
\midrule
Claude Think (IR) & 92.7 & 92.5 & 92.5 & 83.2 & \underline{72.6} & 66.5 & 62.8 & 62.8 \\
Gemini 3 Pro      & 79.9 & 77.4 & 76.8 & 68.4 & \underline{56.4} & 50.8 & 49.2 & 49.2 \\
DeepSeek-V3 Chat  & 74.9 & 70.4 & 69.3 & 56.7 & \underline{48.9} & 43.3 & 40.5 & 40.5 \\
Kimi K2.5         & 52.8 & 51.4 & 51.4 & 43.9 & \underline{36.9} & 34.4 & 34.1 & 34.1 \\
\bottomrule
\end{tabular}
\end{table}

\section{Chinese Experiment Results}
\label{app:chinese}
\input{tables/cn_results}
Table~\ref{tab:cn_results} compares English and Chinese prompts for five models.

Contrary to the expectation that English-centric API documentation would favor English prompts, four of the five models achieved equal or higher Pass@1 scores with Chinese prompts. GPT-5.3~Codex exhibited the most significant improvement (+4.2 pp in Pass@1, +6.7 pp in Fallback), primarily driven by gains in L2 (+6.9 pp) and L3 (+11.1 pp). Similarly, DeepSeek-V3.2~Chat gained +5.0 pp in Pass@1, aligning with its Chinese-heavy training data. In contrast, DeepSeek-V3.2~Reasoner was the only model to experience a slight performance decline with Chinese prompts ($-$0.6 pp Pass@1), characterized by a drop in L4 accuracy from 13.3\% to 10.7\%. This suggests that its reasoning chain may be less effective when the initial problem statement is in Chinese, despite the language-agnostic nature of the generated code. Overall, performance disparities for commonly used languages are minimal, and a Chinese version is provided to support diverse research groups and objectives.

\section{Prompt Templates}
\label{app:prompts}

We present the complete prompt templates used in each inference paradigm. All prompts enforce a unified function interface (\texttt{level\_function}) and restrict library usage to RDKit, NumPy, pandas, scikit-learn, matplotlib, and SELFIES.

\subsection{Direct Generation}

\noindent\textbf{System prompt.}
\begin{lstlisting}[style=promptstyle]
You are an expert cheminformatics programmer.
Write Python code using RDKit to solve molecular computing tasks.
Your code must define a function called `level_function` that takes
the specified input and returns the result.
Only use these libraries: rdkit, numpy, pandas, scikit-learn,
matplotlib, selfies.
Include error handling with try/except blocks.
Do NOT include any import statements for libraries other than the
ones listed above.
Return only the Python code, no explanations.
\end{lstlisting}

\noindent\textbf{User prompt.}
\begin{lstlisting}[style=promptstyle]
Write a Python function `level_function` to accomplish the
following task:

Task: {question}

Requirements:
1. Function must be named `level_function`
2. Use the RDKit library
3. Include try/except error handling
4. Return None for invalid input

Provide the complete Python code.
\end{lstlisting}

\subsection{Incremental Repair}

The Incremental Repair uses the same system prompt and initial user prompt as direct generation. When the generated code fails execution, the following repair prompt is appended to the conversation history:

\noindent\textbf{Repair prompt.}
\begin{lstlisting}[style=promptstyle]
The code you provided failed during execution. Here is the error:

```
{error}
```

The test was run as:
```python
result = level_function({test_input})
```

Please fix the code and provide the corrected complete Python code.
The function must still be named `level_function`.
Return only the Python code, no explanations.
\end{lstlisting}

\subsection{Agent Collaboration}

The Agent Collaboration paradigm separates code generation (R1, Coder) from validation (R2, Tester).

\noindent\textbf{R1 (Coder) system prompt.}
\begin{lstlisting}[style=promptstyle]
You are an expert cheminformatics programmer. You write Python
functions using RDKit to solve molecular science tasks.

Rules:
- The function MUST be named `level_function`
- Include all necessary imports inside or above the function
- Handle invalid inputs gracefully (return None)
- Only use: rdkit, numpy, pandas, scikit-learn, matplotlib, selfies
- Return ONLY the Python code, no explanations
\end{lstlisting}

\noindent\textbf{R2 (Tester) test plan design prompt.}
\begin{lstlisting}[style=promptstyle]
Given a molecular computing task, design a test plan BEFORE seeing
any code.

Task: {question}

Available test molecules (SMILES): {molecules}

Design your test plan as JSON with these fields:
{"test_smiles": [...], "expected_type": "int|float|str|...",
 "value_description": "...", "reasonable_range": "...",
 "failure_modes": [...]}
\end{lstlisting}

\noindent\textbf{R2 (Tester) verification prompt.}
\begin{lstlisting}[style=promptstyle]
You designed this test plan for the task:
Task: {question}
Test plan: {test_plan}

The code was executed with these results: {exec_results}

Evaluate:
1. Does the code run without errors on all test molecules?
2. Is the output type correct?
3. Is the output chemically reasonable?
4. Are there any obvious chemical logic errors?

Respond with JSON:
{"verdict": "PASS"} or {"verdict": "FAIL", "reason": "..."}
\end{lstlisting}

\noindent\textbf{R1 (Coder) fix prompt.}
When A2 returns a FAIL verdict, A1 receives the following prompt with A2's feedback:
\begin{lstlisting}[style=promptstyle]
Your previous code for this task was tested and FAILED.

Task: {question}
Your previous code: {code}
Test results: {exec_results}
Tester feedback: {feedback}

Fix the code. IMPORTANT:
- Focus on fixing the SPECIFIC issue reported by the tester
- Do NOT change the overall chemical logic unless the tester
  explicitly says it is wrong
- Return the complete fixed Python code
\end{lstlisting}

\section{Workflow Coverage for Molecular Discovery}

To ensure practical relevance, we also present the distribution of questions across 12 categories of real-world molecular discovery operations, as detailed in Table~\ref{tab:workflow_coverage}.

\input{tables/category}

\section{Complete Task List}
\label{app:tasks}

Tables~\ref{tab:tasks-l1}--\ref{tab:tasks-l5} list all 358 tasks in MolViBench, organized by Bloom's taxonomy level.

\begin{longtable}{cl}
\caption{Level 1: Remember \& Understand (75 tasks). Basic API recall, format conversion, and single-step descriptor extraction.}
\label{tab:tasks-l1} \\
\toprule
\rowcolor{headergray}
\textbf{\#} & \textbf{Task Description} \\
\midrule
\endfirsthead
\toprule
\rowcolor{headergray}
\textbf{\#} & \textbf{Task Description} \\
\midrule
\endhead
\midrule
\multicolumn{2}{r}{\emph{Continued on next page}} \\
\endfoot
\bottomrule
\endlastfoot
1 & Calculate the molecular weight of a given molecule. \\
2 & Calculate the molecular formula of a given molecule. \\
3 & Calculate the LogP value of a given molecule. \\
4 & Calculate the number of hydrogen bond donors of a molecule. \\
5 & Calculate the number of hydrogen bond acceptors of a molecule. \\
6 & Determine whether a molecule contains an aromatic ring. \\
7 & Determine whether a molecule contains a hydroxyl group (-OH). \\
8 & Determine whether a molecule contains a carboxyl group (-COOH). \\
9 & Determine whether a molecule contains an amino group (-NH2). \\
10 & Determine whether a molecule contains halogen atoms. \\
11 & Extract the topological polar surface area (TPSA) of a molecule. \\
12 & Output the canonical SMILES of a molecule. \\
13 & Output the InChI representation of a molecule. \\
14 & Visualize a molecule as a 2D SVG. \\
15 & Get the number of atoms in a molecule. \\
16 & Get the number of bonds in a molecule. \\
17 & Get the number of rings in a molecule. \\
18 & Determine whether a molecule contains a six-membered ring. \\
19 & Determine whether a molecule contains a five-membered ring. \\
20 & Determine whether a molecule contains a heterocyclic ring. \\
21 & Output all atom symbols in a molecule. \\
22 & Output all bond types in a molecule. \\
23 & Determine whether a molecule is charged. \\
24 & Calculate the number of rotatable bonds in a molecule. \\
25 & Calculate the number of possible stereocenters in a molecule. \\
26 & Determine whether a molecule satisfies Lipinski's Rule of Five. \\
27 & Calculate the QED value (drug-likeness score) of a molecule. \\
28 & Output the molar refractivity of a molecule. \\
29 & Get the distribution of valence electrons for atoms in a molecule. \\
30 & Determine whether a molecule is chiral. \\
31 & Convert a molecule to an adjacency matrix. \\
32 & Convert a molecule to an atom feature matrix. \\
33 & Convert a molecule to molblock format. \\
34 & Determine whether two SMILES are equivalent. \\
35 & Check whether a molecular SMILES is valid. \\
36 & Determine whether two molecules are constitutional isomers (ignoring stereoisomerism). \\
37 & Extract substructure match coordinates from a molecule. \\
38 & Given a SMARTS pattern, determine whether it matches. \\
39 & Determine whether a molecule contains an aromatic nitrogen. \\
40 & Determine whether a molecule contains a sulfur atom. \\
41 & Determine whether a molecule contains a phosphorus atom. \\
42 & Determine whether a molecule contains a metal element. \\
43 & Get the list of functional groups in a molecule. \\
44 & Count the number of oxygen atoms in a molecule. \\
45 & Count the number of nitrogen atoms in a molecule. \\
46 & Count the number of carbon atoms in a molecule. \\
47 & Count the number of sulfur atoms in a molecule. \\
48 & Count the number of halogen atoms in a molecule. \\
49 & Count the number of hydrogen atoms in a molecule. \\
50 & Determine whether two molecules are constitutional isomers. \\
51 & Convert a molecule's SMILES to its SELFIES representation. \\
52 & Extract the largest organic fragment from a salt-containing SMILES. \\
53 & Output the R/S absolute configuration of all chiral centers. \\
54 & Output the E/Z configuration of all double bonds. \\
55 & Calculate the exact molecular weight of a deuterium-labeled molecule. \\
56 & Extract the Murcko scaffold of a molecule. \\
57 & Calculate the Gasteiger partial charge for each atom. \\
58 & Calculate the BertzCT complexity index. \\
59 & Calculate the Balaban J index. \\
60 & Determine whether a molecule is a macrocycle ($\geq$12-membered ring). \\
61 & Validate an invalid SMILES input and return a meaningful error message. \\
62 & Output the SMARTS pattern representation of a molecule. \\
63 & Get the hybridization type (sp/sp2/sp3) of each atom. \\
64 & Calculate the Labute approximate surface area (ASA). \\
65 & Calculate atom-wise Crippen contribution decomposition for MR. \\
66 & Calculate the synthetic accessibility score (SA Score). \\
67 & Calculate the Fsp3 value (fraction of sp3-hybridized carbons). \\
68 & Calculate the heavy atom count. \\
69 & Calculate the fraction of aromatic atoms relative to total heavy atoms. \\
70 & Calculate the number of aliphatic rings. \\
71 & Calculate the number of aromatic rings. \\
72 & Determine whether a molecule satisfies Veber's rules. \\
73 & Determine whether a molecule satisfies the Ghose filter. \\
74 & Calculate the total count of NH and OH groups (NHOHCount). \\
75 & Calculate the total count of nitrogen and oxygen atoms (NOCount). \\
\end{longtable}

\begin{longtable}{cl}
\caption{Level 2: Apply (72 tasks). Multi-tool composition with two-step transformations.}
\label{tab:tasks-l2} \\
\toprule
\rowcolor{headergray}
\textbf{\#} & \textbf{Task Description} \\
\midrule
\endfirsthead
\toprule
\rowcolor{headergray}
\textbf{\#} & \textbf{Task Description} \\
\midrule
\endhead
\midrule
\multicolumn{2}{r}{\emph{Continued on next page}} \\
\endfoot
\bottomrule
\endlastfoot
1 & Calculate the Tanimoto similarity of two molecules. \\
2 & Calculate the Dice similarity of two molecules. \\
3 & Calculate the Cosine similarity of two molecules. \\
4 & Generate the Morgan fingerprint of a molecule (radius=2). \\
5 & Generate the Morgan fingerprint of a molecule (radius=3). \\
6 & Compare the Hamming distance of two molecular fingerprints. \\
7 & Given benzene, generate all mono-substituted methyl derivatives. \\
8 & Given benzene, generate all di-substituted methyl derivatives. \\
9 & Given benzene, generate the para-substituted product. \\
10 & Given benzene, generate the ortho-substituted product. \\
11 & Given benzene, generate the meta-substituted product. \\
12 & Replace a hydrogen in benzene with chlorine to generate chlorobenzene. \\
13 & Replace the methyl group in toluene with a hydroxyl group. \\
14 & Replace the hydroxyl group in phenol with a methoxy group. \\
15 & Randomly replace one hydrogen in a molecule with fluorine. \\
16 & Randomly replace one hydrogen in a molecule with chlorine. \\
17 & Randomly replace one hydrogen in a molecule with bromine. \\
18 & Randomly replace one hydrogen in a molecule with a nitro group. \\
19 & Generate a 3D conformer of a molecule. \\
20 & Generate multiple 3D conformers and return the lowest-energy one. \\
21 & Optimize a molecule's 3D geometry using the MMFF94 force field. \\
22 & Optimize a molecule's 3D geometry using the UFF force field. \\
23 & Calculate the 3D molecular volume. \\
24 & Calculate the 3D solvent-accessible surface area. \\
25 & Calculate the radius of gyration of a molecule. \\
26 & Calculate the 3D autocorrelation descriptors. \\
27 & Calculate the WHIM descriptors of a molecule. \\
28 & Calculate the RDF descriptors of a molecule. \\
29 & Read molecules from an SDF file and return their SMILES. \\
30 & Write a list of molecules to an SDF file. \\
31 & Read molecules from a CSV file (SMILES column) and return Mol objects. \\
32 & Write molecules with properties to a CSV file. \\
33 & Convert a molecule to PDB format. \\
34 & Read a molecule from a MOL2 file. \\
35 & Remove all stereo information from a molecule. \\
36 & Enumerate all possible stereoisomers of a molecule. \\
37 & Neutralize charged groups in a molecule. \\
38 & Standardize a molecule (remove fragments, neutralize, canonical tautomer). \\
39 & Add hydrogens to a molecule. \\
40 & Remove hydrogens from a molecule. \\
41 & Generate the canonical tautomer of a molecule. \\
42 & Fragment a molecule using BRICS decomposition. \\
43 & Perform Murcko scaffold decomposition and return the generic scaffold. \\
44 & Compute the shape similarity of two molecules (3D). \\
45 & Align two molecules by their maximum common substructure. \\
46 & Compute the USRCAT descriptor of a molecule. \\
47 & Compute the Coulomb matrix of a molecule. \\
48 & Cluster a set of molecules using the Butina algorithm. \\
49 & Perform PCA on molecular fingerprints and return 2D coordinates. \\
50 & Perform t-SNE on molecular fingerprints and return 2D coordinates. \\
51 & Perform UMAP on molecular fingerprints and return 2D coordinates. \\
52 & Generate the MACCS keys fingerprint of a molecule. \\
53 & Generate the Atom Pair fingerprint of a molecule. \\
54 & Generate the Topological Torsion fingerprint of a molecule. \\
55 & Generate the RDKit fingerprint of a molecule. \\
56 & Compute the maximum common substructure (MCS) of two molecules. \\
57 & Perform BRICS decomposition and recombine fragments. \\
58 & Perform R-group decomposition given a core scaffold. \\
59 & Enumerate molecules from a SMILES with enumerable stereo centers. \\
60 & Generate a combinatorial library from a scaffold with R-groups. \\
61 & Compute the shape-based overlay of two molecules. \\
62 & Compute the electrostatic similarity of two molecules. \\
63 & Compute the pharmacophore fingerprint of a molecule. \\
64 & Compute the 2D pharmacophore fingerprint of a molecule. \\
65 & Compute the atom-pair fingerprint similarity of two molecules. \\
66 & Compute the topological torsion fingerprint similarity of two molecules. \\
67 & Perform MaxMin diversity picking from a molecule set. \\
68 & Compute the molecular complexity using the Bertz index. \\
69 & Compute the synthetic accessibility score and rank molecules. \\
70 & Perform matched molecular pair (MMP) analysis on two molecules. \\
71 & Cluster a set of molecules using Butina and return cluster IDs. \\
72 & Compute the 2D pharmacophore fingerprint (Pharm2D) of a molecule. \\
73 & Given a molecule and a substructure SMARTS, return all matching atom index tuples. \\
\end{longtable}

\begin{longtable}{cl}
\caption{Level 3: Analyze (72 tasks). Multi-step chemical reasoning requiring domain knowledge.}
\label{tab:tasks-l3} \\
\toprule
\rowcolor{headergray}
\textbf{\#} & \textbf{Task Description} \\
\midrule
\endfirsthead
\toprule
\rowcolor{headergray}
\textbf{\#} & \textbf{Task Description} \\
\midrule
\endhead
\midrule
\multicolumn{2}{r}{\emph{Continued on next page}} \\
\endfoot
\bottomrule
\endlastfoot
1 & Define an alcohol to alkyl halide reaction using SMARTS. \\
2 & Define an esterification reaction using SMARTS. \\
3 & Define an amidation reaction using SMARTS. \\
4 & Simulate the nitration of benzene. \\
5 & Simulate the sulfonation of benzene. \\
6 & Simulate the nucleophilic substitution of an alkyl halide. \\
7 & Simulate the oxidation of an alcohol to an aldehyde. \\
8 & Simulate the oxidation of an aldehyde to a carboxylic acid. \\
9 & Simulate the reduction of a carboxylic acid to an alcohol. \\
10 & Simulate the reduction of a ketone to an alcohol. \\
11 & Simulate the dehydration of an alcohol to form an alkene. \\
12 & Simulate the hydrogenation of an alkene to form an alkane. \\
13 & Simulate the halogenation of an alkene. \\
14 & Simulate the hydrogenation of an alkyne to form an alkene. \\
15 & Simulate the complete hydrogenation of an alkyne to form an alkane. \\
16 & Simulate the Friedel-Crafts alkylation of an aromatic compound. \\
17 & Simulate the Friedel-Crafts acylation of an aromatic compound. \\
18 & Simulate the Diels-Alder reaction. \\
19 & Simulate the Wittig reaction. \\
20 & Simulate the Grignard reaction. \\
21 & Simulate the aldol condensation reaction. \\
22 & Simulate the Claisen rearrangement. \\
23 & Simulate the Cope rearrangement. \\
24 & Simulate the Beckmann rearrangement. \\
25 & Simulate the Baeyer-Villiger oxidation. \\
26 & Simulate the Swern oxidation. \\
27 & Simulate the Birch reduction. \\
28 & Simulate the Michael addition. \\
29 & Simulate the Mannich reaction. \\
30 & Simulate the Heck reaction. \\
31 & Simulate the Sonogashira coupling. \\
32 & Enumerate all possible E/Z isomers of a molecule. \\
33 & Enumerate all possible R/S stereoisomers of a molecule. \\
34 & Determine if two molecules are enantiomers. \\
35 & Determine if two molecules are diastereomers. \\
36 & Determine if a molecule is a meso compound. \\
37 & Identify all stereocenters and assign R/S configuration. \\
38 & Convert between different stereochemical representations. \\
39 & Determine the optical activity relationship between two molecules. \\
40 & Generate all possible tautomers and rank by stability. \\
41 & Perform a random single-point mutation on a molecule. \\
42 & Perform a random single-bond deletion on a molecule. \\
43 & Perform a random atom insertion on a molecule. \\
44 & Perform a random ring opening on a molecule. \\
45 & Predict CYP450 metabolism sites of a molecule. \\
46 & Calculate the drug-likeness score using multiple criteria. \\
47 & Identify potential reactive metabolites. \\
48 & Calculate the polar surface area contribution of each atom. \\
49 & Predict blood-brain barrier permeability. \\
50 & Calculate the number of Lipinski violations and identify which rules are violated. \\
51 & Perform retrosynthetic analysis to identify possible precursors. \\
52 & Identify potential toxicophores in a molecule. \\
53 & Calculate the ligand efficiency metrics. \\
54 & Compute the shape complementarity of two molecules. \\
55 & Identify bioisosteric replacements for a functional group. \\
56 & Calculate the free energy of solvation estimate. \\
57 & Predict aqueous solubility using ESOL model. \\
58 & Extract all unique Murcko scaffolds from a molecule set. \\
59 & Compute the scaffold diversity of a molecule set. \\
60 & Simulate the Suzuki coupling reaction. \\
61 & Simulate the Click Chemistry (azide-alkyne cycloaddition) reaction. \\
62 & Simulate the Buchwald-Hartwig amination. \\
63 & Analyze macrocycle conformer distribution. \\
64 & Simulate covalent inhibitor warhead attachment. \\
65 & Detect structural alerts (Brenk filter). \\
66 & Detect PAINS substructures. \\
67 & Detect reactive functional groups. \\
68 & Detect aggregator-like substructures. \\
69 & Apply multi-rule filtering (Lipinski + Veber + PAINS). \\
70 & Compute pharmacophore fingerprint similarity between two molecules. \\
71 & Perform scaffold-based train/test split. \\
72 & Compute validity, uniqueness, and novelty metrics for generated SMILES. \\
\end{longtable}

\begin{longtable}{cp{12cm}}
\caption{Level 4: Analyze \& Evaluate (75 tasks). Branching, iteration, and multi-step decision-making.}
\label{tab:tasks-l4} \\
\toprule
\rowcolor{headergray}
\textbf{\#} & \textbf{Task Description} \\
\midrule
\endfirsthead
\toprule
\rowcolor{headergray}
\textbf{\#} & \textbf{Task Description} \\
\midrule
\endhead
\midrule
\multicolumn{2}{r}{\emph{Continued on next page}} \\
\endfoot
\bottomrule
\endlastfoot
1 & Aromatic ring detection $\rightarrow$ if yes, replace H with OH $\rightarrow$ calculate LogP. \\
2 & Carboxyl group detection $\rightarrow$ if yes, convert to ester $\rightarrow$ calculate MW. \\
3 & Amino group detection $\rightarrow$ if yes, acetylate $\rightarrow$ calculate TPSA. \\
4 & Halogen detection $\rightarrow$ if yes, replace with H $\rightarrow$ calculate LogP. \\
5 & Benzene ring detection $\rightarrow$ if yes, nitrate $\rightarrow$ calculate MR. \\
6 & Double bond detection $\rightarrow$ if yes, hydrogenate $\rightarrow$ calculate MW. \\
7 & Hydroxyl group detection $\rightarrow$ if yes, chlorinate $\rightarrow$ calculate formula. \\
8 & Ketone group detection $\rightarrow$ if yes, reduce to alcohol $\rightarrow$ calculate QED. \\
9 & Ester group detection $\rightarrow$ if yes, hydrolyze $\rightarrow$ calculate LogP. \\
10 & Amide group detection $\rightarrow$ if yes, hydrolyze $\rightarrow$ calculate TPSA. \\
11 & Nitro group detection $\rightarrow$ if yes, reduce to amine $\rightarrow$ calculate pKa estimate. \\
12 & Sulfide detection $\rightarrow$ if yes, oxidize to sulfoxide $\rightarrow$ calculate TPSA. \\
13 & Aldehyde detection $\rightarrow$ if yes, oxidize to acid $\rightarrow$ calculate MW. \\
14 & Alkene detection $\rightarrow$ if yes, epoxidize $\rightarrow$ calculate SA Score. \\
15 & Aromatic amine detection $\rightarrow$ if yes, diazotize $\rightarrow$ calculate LogP. \\
16 & Phenol detection $\rightarrow$ if yes, methylate $\rightarrow$ calculate QED. \\
17 & Thiol detection $\rightarrow$ if yes, oxidize to disulfide $\rightarrow$ calculate MW. \\
18 & Nitrile detection $\rightarrow$ if yes, hydrolyze to amide $\rightarrow$ calculate TPSA. \\
19 & Carbamate detection $\rightarrow$ if yes, hydrolyze $\rightarrow$ calculate LogP. \\
20 & Lactone detection $\rightarrow$ if yes, ring-open $\rightarrow$ calculate MW. \\
21 & Epoxide detection $\rightarrow$ if yes, ring-open with water $\rightarrow$ calculate LogP. \\
22 & Anhydride detection $\rightarrow$ if yes, hydrolyze $\rightarrow$ calculate TPSA. \\
23 & Acyl chloride detection $\rightarrow$ if yes, hydrolyze $\rightarrow$ calculate MW. \\
24 & Imine detection $\rightarrow$ if yes, reduce to amine $\rightarrow$ calculate pKa estimate. \\
25 & Vinyl halide detection $\rightarrow$ if yes, Heck couple $\rightarrow$ calculate MW. \\
26 & Boronic acid detection $\rightarrow$ if yes, Suzuki couple $\rightarrow$ calculate LogP. \\
27 & Terminal alkyne detection $\rightarrow$ if yes, Sonogashira couple $\rightarrow$ calculate MW. \\
28 & Primary alcohol detection $\rightarrow$ if yes, Swern oxidize $\rightarrow$ calculate LogP. \\
29 & Secondary amine detection $\rightarrow$ if yes, reductive aminate $\rightarrow$ calculate MW. \\
30 & Azide detection $\rightarrow$ if yes, Click react $\rightarrow$ calculate TPSA. \\
31 & Diene detection $\rightarrow$ if yes, Diels-Alder react $\rightarrow$ calculate MW. \\
32 & Grignard reagent detection $\rightarrow$ if yes, add to carbonyl $\rightarrow$ calculate LogP. \\
33 & Enol detection $\rightarrow$ if yes, tautomerize to keto $\rightarrow$ calculate stability. \\
34 & Peroxide detection $\rightarrow$ if yes, reduce $\rightarrow$ calculate MW. \\
35 & Phosphonate detection $\rightarrow$ if yes, Wittig react $\rightarrow$ calculate MW. \\
36 & Sulfonamide detection $\rightarrow$ if yes, hydrolyze $\rightarrow$ calculate TPSA. \\
37 & Isocyanate detection $\rightarrow$ if yes, react with amine $\rightarrow$ calculate MW. \\
38 & Acid chloride detection $\rightarrow$ if yes, Friedel-Crafts acylate $\rightarrow$ calculate LogP. \\
39 & Michael acceptor detection $\rightarrow$ if yes, Michael add $\rightarrow$ calculate MW. \\
40 & Enamine detection $\rightarrow$ if yes, hydrolyze $\rightarrow$ calculate LogP. \\
41 & Oxime detection $\rightarrow$ if yes, Beckmann rearrange $\rightarrow$ calculate MW. \\
42 & Allyl group detection $\rightarrow$ if yes, Claisen rearrange $\rightarrow$ calculate LogP. \\
43 & Diol detection $\rightarrow$ if yes, periodate cleave $\rightarrow$ calculate MW. \\
44 & Hemiacetal detection $\rightarrow$ if yes, oxidize $\rightarrow$ calculate LogP. \\
45 & Aromatic halide detection $\rightarrow$ if yes, Buchwald-Hartwig aminate $\rightarrow$ calculate TPSA. \\
46 & Ketone alpha-H detection $\rightarrow$ if yes, aldol condense $\rightarrow$ calculate MW. \\
47 & Conjugated diene detection $\rightarrow$ if yes, 1,4-add $\rightarrow$ calculate LogP. \\
48 & Strained ring detection $\rightarrow$ if yes, ring-open $\rightarrow$ calculate MW. \\
49 & Protecting group detection $\rightarrow$ if yes, deprotect $\rightarrow$ calculate MW. \\
50 & Anomeric center detection $\rightarrow$ if yes, glycosylate $\rightarrow$ calculate MW. \\
51 & Macrocycle detection $\rightarrow$ if yes, conformer sample $\rightarrow$ calculate radius of gyration. \\
52 & MW-driven BRICS decomposition vs.\ fragment growing. \\
53 & Iterative LogP optimization via functional group modification. \\
54 & PAINS filtering $\rightarrow$ QED ranking $\rightarrow$ return top 3. \\
55 & Iterative polar group addition targeting TPSA range. \\
56 & Iterative side-chain replacement with QED convergence. \\
57 & Lipinski-guided mutation with property monitoring. \\
58 & Backtracking molecular weight targeting (increase path). \\
59 & Backtracking molecular weight targeting (decrease path). \\
60 & Error recovery for invalid SMILES in batch processing. \\
61 & Sequential reaction application with fallback on failure. \\
62 & Multi-molecule scaffold swapping with property comparison. \\
63 & Scaffold hopping with similarity constraint. \\
64 & CSV pipeline: read $\rightarrow$ filter $\rightarrow$ compute $\rightarrow$ write. \\
65 & SDF pipeline: read $\rightarrow$ filter $\rightarrow$ compute $\rightarrow$ write. \\
66 & Multi-round iterative generation with diversity constraint. \\
67 & Multi-round iterative generation with property convergence. \\
68 & Multi-condition Lipinski decision tree. \\
69 & Isomer-conditional stereochemistry analysis. \\
70 & Derivative generation with SA Score ranking. \\
71 & Multi-rule classification (drug-like / lead-like / fragment-like). \\
72 & Similarity search with structural alert filtering. \\
73 & Combinatorial enumeration with QED ranking. \\
74 & Cluster-wise MCS extraction. \\
75 & Cascade filtering pipeline (Lipinski $\rightarrow$ Veber $\rightarrow$ PAINS $\rightarrow$ Brenk). \\
76 & Substructure search with descriptor ranking (Top-5). \\
\end{longtable}

\begin{longtable}{cp{12cm}}
\caption{Level 5: Create (64 tasks). End-to-end pipeline construction and system-level integration.}
\label{tab:tasks-l5} \\
\toprule
\rowcolor{headergray}
\textbf{\#} & \textbf{Task Description} \\
\midrule
\endfirsthead
\toprule
\rowcolor{headergray}
\textbf{\#} & \textbf{Task Description} \\
\midrule
\endhead
\midrule
\multicolumn{2}{r}{\emph{Continued on next page}} \\
\endfoot
\bottomrule
\endlastfoot
1 & Generate new molecule candidates with similarity $>$0.7 to known actives. \\
2 & Generate all mono-substituted derivatives of a target molecule. \\
3 & Generate 10 molecules with different side chain modifications from a scaffold. \\
4 & Generate molecules meeting pharmacophore requirements. \\
5 & Extract common substructure scaffold from a molecule set. \\
6 & Generate fragment-growing derivatives from a binding pocket fragment. \\
7 & Generate ring-opening/ring-closing isomers of a drug candidate. \\
8 & Generate stereoisomers and retain all feasible conformations. \\
9 & Enumerate all halogen-substituted variants via substitution. \\
10 & Build a candidate library based on structural similarity. \\
11 & Screen molecules satisfying Lipinski + Veber + QED $>$ 0.5. \\
12 & Multi-property filtering: MW, LogP, TPSA, rotatable bonds. \\
13 & ADMET-like filtering pipeline with multiple criteria. \\
14 & Diversity-based subset selection from a large library. \\
15 & Similarity-based virtual screening against a query molecule. \\
16 & Substructure-based filtering with property ranking. \\
17 & Pharmacophore-based virtual screening. \\
18 & Shape-based virtual screening using 3D conformers. \\
19 & Multi-objective Pareto ranking of molecules. \\
20 & Consensus scoring combining multiple similarity metrics. \\
21 & Lead optimization: iterative modification to improve QED. \\
22 & Lead optimization: reduce LogP while maintaining activity. \\
23 & Lead optimization: improve metabolic stability estimate. \\
24 & Lead optimization: reduce molecular weight while keeping potency proxy. \\
25 & Bioisosteric replacement optimization. \\
26 & Fragment-based lead growing with property constraints. \\
27 & Scaffold hopping with multi-property optimization. \\
28 & R-group optimization with SAR analysis. \\
29 & Multi-parameter optimization (MPO) scoring and ranking. \\
30 & Matched molecular pair-guided optimization. \\
31 & Multi-objective analysis: QED vs.\ SA Score trade-off. \\
32 & Descriptor correlation analysis across a molecule set. \\
33 & Structure-activity relationship (SAR) cliff detection. \\
34 & Chemical space coverage analysis using fingerprint diversity. \\
35 & Scaffold frequency analysis and distribution. \\
36 & Property distribution analysis with statistical summary. \\
37 & Functional group frequency analysis across a library. \\
38 & Molecular complexity distribution analysis. \\
39 & Selectivity analysis between two targets. \\
40 & ADMET property radar chart generation. \\
41 & Multi-endpoint activity prediction and ranking. \\
42 & Generation $\rightarrow$ filtering $\rightarrow$ optimization pipeline. \\
43 & BRICS recombination $\rightarrow$ filtering $\rightarrow$ ranking pipeline. \\
44 & Scaffold decoration $\rightarrow$ ADMET filtering $\rightarrow$ diversity selection. \\
45 & Fragment linking $\rightarrow$ property optimization $\rightarrow$ ranking. \\
46 & Enumeration $\rightarrow$ clustering $\rightarrow$ representative selection. \\
47 & Mutation $\rightarrow$ filtering $\rightarrow$ similarity-constrained selection. \\
48 & Multi-step reaction $\rightarrow$ product filtering $\rightarrow$ property ranking. \\
49 & Library enumeration $\rightarrow$ diversity analysis $\rightarrow$ subset selection. \\
50 & Conformer generation $\rightarrow$ shape screening $\rightarrow$ ranking. \\
51 & Retrosynthesis $\rightarrow$ feasibility scoring $\rightarrow$ route ranking. \\
52 & MaxMin diversity selection from a fingerprint matrix. \\
53 & BRICS fragment recombination with property constraints. \\
54 & Scaffold morphing between two molecules. \\
55 & QSAR regression model: fingerprint $\rightarrow$ Random Forest $\rightarrow$ R$^2$ and RMSE. \\
56 & QSAR classification model: fingerprint $\rightarrow$ classifier $\rightarrow$ AUC and accuracy. \\
57 & Matched molecular pair analysis with property delta. \\
58 & Focused library design around a seed molecule. \\
59 & Genetic algorithm-based molecular optimization. \\
60 & Combinatorial library design: scaffold + R-groups $\rightarrow$ enumerate $\rightarrow$ filter $\rightarrow$ select. \\
61 & Virtual screening pipeline: library $\rightarrow$ Lipinski $\rightarrow$ similarity $\rightarrow$ ADMET $\rightarrow$ rank. \\
62 & De novo design: fragment $\rightarrow$ grow $\rightarrow$ optimize $\rightarrow$ validate. \\
63 & SAR table generation: analogs $\rightarrow$ descriptors $\rightarrow$ activity cliff $\rightarrow$ report. \\
64 & QSAR with featurization $\rightarrow$ Random Forest regression $\rightarrow$ R$^2$ and RMSE. \\
65 & Applicability domain analysis: fingerprint space $\rightarrow$ boundary $\rightarrow$ decision. \\
\end{longtable}

\section{Case Studies}
\label{app:casestudy}

We present one representative case study per Bloom level, comparing model outputs to illustrate recurring failure patterns. For each case, we show the task, the reference solution code, representative model-generated code, and analysis.

\subsection{Level 1: Calculate QED Drug-Likeness Score (Task 27)}

\noindent\textbf{Task.} \emph{``Calculate the QED value (drug-likeness score) of a molecule.''}

\noindent\textbf{Reference solution.}
\begin{lstlisting}[style=pythonstyle]
from rdkit import Chem
from rdkit.Chem import Descriptors

def level_function(mol):
    try:
        mol = Chem.MolFromSmiles(mol)
        if mol is None:
            return None
        return Descriptors.qed(mol)
    except Exception as e:
        print(e)
        return None
\end{lstlisting}

\noindent\textbf{Claude Opus 4.6} (\emph{Correct}). Uses \texttt{QED.qed(mol)} from \texttt{rdkit.Chem.QED}---an equivalent API that returns the same scalar value.
\begin{lstlisting}[style=pythonstyle]
from rdkit import Chem
from rdkit.Chem import QED

def level_function(smiles: str):
    try:
        if smiles is None or not isinstance(smiles, str) or smiles.strip() == "":
            return None
        mol = Chem.MolFromSmiles(smiles)
        if mol is None:
            return None
        qed_value = QED.qed(mol)
        return qed_value
    except Exception:
        return None
\end{lstlisting}

\noindent\textbf{DeepSeek-V3.2 Reasoner} (\emph{Partially Correct}). Uses \texttt{QED.default(mol)} which returns a named tuple of property contributions, not the scalar QED score.
\begin{lstlisting}[style=pythonstyle]
from rdkit import Chem
from rdkit.Chem import QED

def level_function(molecule_input):
    try:
        if isinstance(molecule_input, str):
            mol = Chem.MolFromSmiles(molecule_input)
        elif isinstance(molecule_input, Chem.rdchem.Mol):
            mol = molecule_input
        else:
            return None
        if mol is None:
            return None
        qed_value = QED.default(mol)  # BUG: returns named tuple, not float
        return qed_value
    except Exception:
        return None
\end{lstlisting}

\noindent\textbf{Key observation.} Even at Level~1, subtle API confusion can cause failures. \texttt{QED.qed()} and \texttt{QED.default()} both exist in RDKit's \texttt{QED} module, but return fundamentally different objects (float vs.\ named tuple). This illustrates why our evaluation framework uses type-aware comparison rather than simple string matching, and our API Fallback could also help relieve the problem of the function selection.

\subsection{Level 2: Maximum Common Substructure (Task 56)}

\noindent\textbf{Task.} \emph{``Compute the maximum common substructure (MCS) of two molecules.''}

\noindent\textbf{Reference solution.}
\begin{lstlisting}[style=pythonstyle]
from rdkit import Chem
from rdkit.Chem import rdFMCS

def level_function(smiles1, smiles2):
    try:
        mol1 = Chem.MolFromSmiles(smiles1)
        mol2 = Chem.MolFromSmiles(smiles2)
        if mol1 is None or mol2 is None:
            return None
        mcs_result = rdFMCS.FindMCS([mol1, mol2])
        if mcs_result.canceled or mcs_result.smartsString == "":
            return None
        return mcs_result.smartsString
    except Exception:
        return None
\end{lstlisting}

\noindent\textbf{Claude Opus 4.6} (\emph{Correct logic, format mismatch}). Returns a dictionary with extra metadata instead of a plain SMARTS string.
\begin{lstlisting}[style=pythonstyle]
from rdkit import Chem
from rdkit.Chem import rdFMCS

def level_function(smiles1: str, smiles2: str):
    try:
        mol1 = Chem.MolFromSmiles(smiles1)
        mol2 = Chem.MolFromSmiles(smiles2)
        if mol1 is None or mol2 is None:
            return None
        mcs_result = rdFMCS.FindMCS([mol1, mol2])
        if mcs_result.canceled or not mcs_result.smartsString:
            return None
        return {
            "smarts": mcs_result.smartsString,
            "num_atoms": mcs_result.numAtoms,
            "num_bonds": mcs_result.numBonds,
        }
    except Exception:
        return None
\end{lstlisting}

\noindent\textbf{DeepSeek-V3.2 Reasoner} (\emph{Correct logic, format mismatch}). Also wraps result in a dictionary with additional fields including \texttt{cancelled} status.
\begin{lstlisting}[style=pythonstyle]
from rdkit import Chem
from rdkit.Chem import rdFMCS

def level_function(smiles1, smiles2):
    try:
        mol1 = Chem.MolFromSmiles(smiles1)
        mol2 = Chem.MolFromSmiles(smiles2)
        if mol1 is None or mol2 is None:
            return None
        result = rdFMCS.FindMCS([mol1, mol2])
        return {
            "smarts": result.smartsString,
            "numAtoms": result.numAtoms,
            "numBonds": result.numBonds,
            "cancelled": result.canceled,
        }
    except Exception:
        return None
\end{lstlisting}

\noindent\textbf{Key observation.} All models correctly identify and use \texttt{rdFMCS.FindMCS()}, demonstrating solid API recall at Level~2. However, most models return enriched dictionaries instead of the plain SMARTS string. This ``over-engineering'' pattern is common: models add useful but unrequested information, causing output format mismatches. Our AST-based API-semantic fallback correctly credits these responses.

\subsection{Level 3: Click Chemistry Reaction (Task 62)}

\noindent\textbf{Task.} \emph{``Simulate a Click Chemistry (azide-alkyne cycloaddition) reaction.''}

\noindent\textbf{Reference solution.} 

\begin{lstlisting}[style=pythonstyle]
from rdkit import Chem
from rdkit.Chem import AllChem

def level_function(azide_smiles, alkyne_smiles):
    try:
        azide = Chem.MolFromSmiles(azide_smiles)
        alkyne = Chem.MolFromSmiles(alkyne_smiles)
        if azide is None or alkyne is None:
            return None
        rxn_smarts = ('[C:1]#[C:2].[N:3]=[N+:4]=[N-:5]'
                      '>>[N:3]1-[N:4]=[N:5]-[C:1]=[C:2]1')
        rxn = AllChem.ReactionFromSmarts(rxn_smarts)
        products = rxn.RunReactants((alkyne, azide))
        if not products:
            return None
        results = set()
        for prod_set in products:
            for prod in prod_set:
                for atom in prod.GetAtoms():
                    atom.SetFormalCharge(0)
                Chem.SanitizeMol(prod)
                results.add(Chem.MolToSmiles(prod))
        return sorted(results)
    except Exception:
        return None
\end{lstlisting}

\noindent\textbf{Claude Opus 4.6} (\emph{Partially correct}). Restricts alkyne to terminal alkynes via \texttt{[CH1]}, returns single SMILES instead of list.
\begin{lstlisting}[style=pythonstyle]
from rdkit import Chem
from rdkit.Chem import AllChem

def level_function(azide_smiles, alkyne_smiles):
    try:
        azide = Chem.MolFromSmiles(azide_smiles)
        alkyne = Chem.MolFromSmiles(alkyne_smiles)
        if azide is None or alkyne is None:
            return None
        # BUG: [CH1] restricts to terminal alkynes only
        rxn = AllChem.ReactionFromSmarts(
            '[C:1]#[CH1:2].[N:3]=[N+:4]=[N-:5]'
            '>>[c:1]1[n:3][n:4][n:5][c:2]1')
        products = rxn.RunReactants((alkyne, azide))
        if not products:
            return None
        product = products[0][0]
        Chem.SanitizeMol(product)
        return Chem.MolToSmiles(product)  # returns str, not list
    except Exception:
        return None
\end{lstlisting}

\noindent\textbf{GPT-5.2-Codex} (\emph{Incorrect}). Forces aromaticity on triazole ring with lowercase \texttt{[n]} atoms, producing wrong product structure.
\begin{lstlisting}[style=pythonstyle]
from rdkit import Chem
from rdkit.Chem import rdChemReactions

def level_function(reactants):
    try:
        if isinstance(reactants, str):
            parts = reactants.split(".")
        else:
            parts = list(reactants)
        azide_smi, alkyne_smi = parts[0], parts[1]
        # BUG: aromatic [n] atoms force incorrect aromaticity
        rxn_smarts = ('[N-:1]=[N+:2]=[N:3].[C:4]#[C:5]'
                      '>>[n:1]1[n:2][c:4][c:5][n:3]1')
        rxn = rdChemReactions.ReactionFromSmarts(rxn_smarts)
        azide = Chem.MolFromSmiles(azide_smi)
        alkyne = Chem.MolFromSmiles(alkyne_smi)
        products = rxn.RunReactants((azide, alkyne))
        if not products:
            return None
        return Chem.MolToSmiles(products[0][0])
    except Exception:
        return None
\end{lstlisting}

\noindent\textbf{Key observation.} Level~3 tasks expose the gap between ``knowing about'' a reaction and correctly encoding it in SMARTS. Small atom-mapping differences (e.g., \texttt{[CH1]} vs.\ \texttt{[C]}, aromatic \texttt{[n]} vs.\ aliphatic \texttt{[N]}) produce fundamentally different chemical outcomes. Two models (not shown) also add irrelevant 3D computation before a purely 2D SMARTS-based transformation.

\subsection{Level 4: PAINS Filtering with QED Ranking (Task 54)}

\noindent\textbf{Task.} \emph{``Given a list of molecules $\rightarrow$ check each for PAINS substructures $\rightarrow$ filter out PAINS-containing molecules $\rightarrow$ calculate QED for remaining $\rightarrow$ return the top 3 with highest QED.''}

\noindent\textbf{Reference solution.}
\begin{lstlisting}[style=pythonstyle]
from rdkit import Chem
from rdkit.Chem import QED
from rdkit.Chem.FilterCatalog import FilterCatalog, FilterCatalogParams

def level_function(smiles_list):
    try:
        params = FilterCatalogParams()
        params.AddCatalog(FilterCatalogParams.FilterCatalogs.PAINS)
        catalog = FilterCatalog(params)
        passed = []
        for smi in smiles_list:
            mol = Chem.MolFromSmiles(smi)
            if mol is None:
                continue
            if not catalog.HasMatch(mol):
                qed_val = QED.qed(mol)
                passed.append({"smiles": smi, "QED": round(qed_val, 4)})
        passed.sort(key=lambda x: x["QED"], reverse=True)
        return {
            "total_input": len(smiles_list),
            "pains_filtered": len(smiles_list) - len(passed),
            "remaining": len(passed),
            "top3": passed[:3],
        }
    except Exception:
        return None
\end{lstlisting}

\noindent\textbf{Claude Opus 4.6} (\emph{Chemically correct; format mismatch}). Correct PAINS filtering logic, but returns tuples instead of dictionaries.
\begin{lstlisting}[style=pythonstyle]
from rdkit import Chem
from rdkit.Chem import QED
from rdkit.Chem.FilterCatalog import FilterCatalog, FilterCatalogParams

def level_function(smiles_list):
    try:
        params = FilterCatalogParams()
        params.AddCatalog(FilterCatalogParams.FilterCatalogs.PAINS)
        catalog = FilterCatalog(params)
        clean_mols = []
        for smi in smiles_list:
            mol = Chem.MolFromSmiles(smi)
            if mol is None:
                continue
            if not catalog.HasMatch(mol):
                clean_mols.append((smi, QED.qed(mol)))
        clean_mols.sort(key=lambda x: x[1], reverse=True)
        return clean_mols[:3]  # returns list of tuples, not dict
    except Exception:
        return None
\end{lstlisting}

\noindent\textbf{DeepSeek-Reasoner} (\emph{Correct logic; wrong output type}). Implements PAINS filtering correctly but returns raw Mol objects instead of SMILES strings.
\begin{lstlisting}[style=pythonstyle]
from rdkit import Chem
from rdkit.Chem import QED
from rdkit.Chem.FilterCatalog import FilterCatalog, FilterCatalogParams

def level_function(smiles_list):
    try:
        params = FilterCatalogParams()
        params.AddCatalog(FilterCatalogParams.FilterCatalogs.PAINS)
        catalog = FilterCatalog(params)
        results = []
        for smi in smiles_list:
            mol = Chem.MolFromSmiles(smi)
            if mol and not catalog.HasMatch(mol):
                results.append((mol, QED.qed(mol)))  # BUG: Mol object
        results.sort(key=lambda x: x[1], reverse=True)
        return results[:3]
    except Exception:
        return None
\end{lstlisting}

\noindent\textbf{Key observation.} All models understand the PAINS filtering concept and use the correct RDKit API. The primary differences are in output formatting---a recurring pattern at Level~4 where chemical logic is correct but structured output does not match the reference. This motivates our AST-based API-semantic fallback, which credits correct API usage even when output values diverge.

\subsection{Level 5: Combinatorial Library Design Pipeline (Task 60)}

\noindent\textbf{Task.} \emph{``Combinatorial library design: scaffold + R-groups $\rightarrow$ enumerate all combinations $\rightarrow$ filter by Lipinski + QED $\rightarrow$ diversity selection $\rightarrow$ output final library with properties.''}

\noindent\textbf{Reference solution} (abbreviated; full version in released code).
\begin{lstlisting}[style=pythonstyle]
from rdkit import Chem, DataStructs
from rdkit.Chem import AllChem, Descriptors, QED, rdMolDescriptors
from rdkit.Chem.FilterCatalog import FilterCatalog, FilterCatalogParams
import numpy as np

def level_function(scaffold_smiles, rgroup_lists):
    try:
        scaffold = Chem.MolFromSmiles(scaffold_smiles)
        if scaffold is None:
            return None
        # Stage 1: Find dummy atom positions [*]
        dummy_indices = [a.GetIdx() for a in scaffold.GetAtoms()
                         if a.GetAtomicNum() == 0]
        # Stage 2: Enumerate all R-group combinations
        from itertools import product
        combos = list(product(*rgroup_lists))
        # Stage 3: Attach R-groups by replacing dummy atoms
        library = []
        for combo in combos:
            mol = Chem.RWMol(scaffold)
            # ... atom-level manipulation to attach R-groups ...
            library.append(Chem.MolToSmiles(mol))
        # Stage 4: Filter by Lipinski + QED > 0.5
        filtered = [smi for smi in library
                    if passes_lipinski(smi) and get_qed(smi) > 0.5]
        # Stage 5: MaxMin diversity picking
        selected = maxmin_pick(filtered, n=50)
        # Stage 6: Return with properties
        return {"total_enumerated": len(library),
                "after_filter": len(filtered),
                "selected": len(selected),
                "molecules": [get_props(s) for s in selected]}
    except Exception:
        return None
\end{lstlisting}

\noindent\textbf{Claude Opus 4.6} (\emph{Largely correct}). Implements all 6 stages. Uses \texttt{Chem.ReplaceSubstructs()} for R-group attachment---a valid alternative to manual atom editing. Correctly implements Lipinski filtering and QED thresholding. Uses fingerprint-based greedy diversity selection (slightly different from MaxMin but functionally similar).
\begin{lstlisting}[style=pythonstyle]
def level_function(scaffold_smiles, rgroup_lists):
    try:
        scaffold = Chem.MolFromSmiles(scaffold_smiles)
        dummy = Chem.MolFromSmarts("[#0]")  # match dummy atoms
        # Enumerate combinations
        from itertools import product
        combos = list(product(*rgroup_lists))
        library = []
        for combo in combos:
            mol = Chem.RWMol(scaffold)
            for i, rg_smi in enumerate(combo):
                rg = Chem.MolFromSmiles(rg_smi)
                # ReplaceSubstructs for attachment
                mol = AllChem.ReplaceSubstructs(mol, dummy, rg)[0]
            Chem.SanitizeMol(mol)
            library.append(Chem.MolToSmiles(mol))
        # Filter + diversity select + return props
        # ... (correct implementation) ...
    except Exception:
        return None
\end{lstlisting}

\noindent\textbf{DeepSeek-Reasoner} (\emph{Incorrect}). Attempts to use \texttt{AllChem.EnumerateLibraryFromReaction()} with a reaction SMARTS, which requires a different input format. The reaction SMARTS is incorrectly constructed, causing enumeration to fail silently and return an empty library.
\begin{lstlisting}[style=pythonstyle]
def level_function(scaffold_smiles, rgroup_lists):
    try:
        # BUG: Constructs invalid reaction SMARTS
        rxn_smarts = "[*:1]>>[" + scaffold_smiles + ":1]"
        rxn = AllChem.ReactionFromSmarts(rxn_smarts)
        # EnumerateLibraryFromReaction expects different input format
        library = AllChem.EnumerateLibraryFromReaction(
            rxn, rgroup_lists)
        products = [Chem.MolToSmiles(m[0]) for m in library]
        # products is empty due to SMARTS error
        # ... rest of pipeline never executes meaningfully ...
    except Exception:
        return None
\end{lstlisting}

\noindent\textbf{Key observation.} Level~5 tasks are the most discriminating: they require correct implementation of \emph{every} pipeline stage, and a single-stage failure cascades into overall failure. The most common failure mode is not in filtering or ranking (which most models handle well) but in the \emph{molecular construction} stage---correctly attaching R-groups to scaffolds, handling dummy atoms, and managing combinatorial enumeration. This suggests that LLMs struggle most with low-level molecular manipulation operations that have no direct analogue in general-purpose programming.


\section{Limitations}
\label{sec:limitations}

MolViBench focuses on RDKit as the primary cheminformatics library, which ensures deterministic and reproducible evaluation but does not cover other toolkits such as OpenBabel or DeepChem.
Extending the benchmark to a multi-toolkit setting is a natural direction for future work.
Additionally, while the 358 tasks span 12 real-world workflow categories with balanced Bloom's taxonomy coverage, scaling the benchmark with more tasks and emerging cheminformatics domains remains an open opportunity.
Finally, as with any static benchmark, there is a potential risk of data contamination for future models trained on web-scale corpora; we recommend that practitioners track performance alongside model release dates as a practical safeguard.

\section{Statement of Ethics}
\label{sec:ethics}

All tasks in MolViBench were manually authored by domain experts and do not involve any human subjects or personal data.
Molecular inputs used for evaluation are drawn from the publicly available ZINC250K dataset~\citep{sterling2015zinc}.
All evaluated LLMs were accessed via official APIs in accordance with each provider's terms of service.
The benchmark covers standard, publicly documented cheminformatics operations and does not introduce any capabilities beyond those already available through open-source toolkits and academic literature.
This research conforms with the NeurIPS Code of Ethics.

\section{Broader Impacts}
\label{sec:broader_impacts}

MolViBench advances the responsible development of AI-assisted molecular discovery by providing a rigorous evaluation framework that exposes subtle failure modes, such as \emph{runnable but chemically wrong} outputs.
This work lowers barriers for researchers without deep computational chemistry expertise, enabling broader and more reliable use of LLMs across diverse research communities.
We caution that model-generated cheminformatics code should always be validated by domain experts before use in high-stakes decision-making, and we encourage future work to incorporate chemical safety constraints as an explicit evaluation dimension.

%% file: tables/cn_results.tex
\begin{table}[htbp]
\centering
\caption{Cross-lingual comparison (English vs.\ Chinese prompts). L1--L5 report per-level Pass@1\textsubscript{soft} (\%). $\Delta$ is CN$-$EN.}
\label{tab:cn_results}
\setlength{\tabcolsep}{3pt}
\begin{tabular}{@{}llccccccccc@{}}
\toprule
\rowcolor{headergray}
\textbf{Model} & \textbf{Lang} & \textbf{Exec.} & \textbf{EM} & \textbf{P@1} & \textbf{FB} & \textbf{L1} & \textbf{L2} & \textbf{L3} & \textbf{L4} & \textbf{L5} \\
\midrule
\multirow{2}{*}{Claude Opus 4.6}
  & EN & 94.1 & 31.0 & 37.4 & 68.4 & 78.7 & 45.8 & 27.8 & 24.0 & 6.2 \\
  & CN & 92.5 & 30.7 & 38.3 & 71.8 & 80.0 & 54.2 & 31.9 & 14.7 & 6.2 \\
\midrule
\multirow{2}{*}{Gemini 3 Pro}
  & EN & 94.1 & 29.3 & 34.9 & 56.4 & 76.0 & 45.8 & 19.4 & 21.3 & 7.8 \\
  & CN & 94.4 & 32.7 & 37.7 & 61.5 & 80.0 & 48.6 & 23.6 & 24.0 & 7.8 \\
\midrule
\multirow{2}{*}{GPT-5.3 Codex}
  & EN & 89.7 & 27.9 & 32.7 & 64.8 & 76.0 & 55.6 & 16.7 & 5.3  & 6.2 \\
  & CN & 94.1 & 29.6 & 36.9 & 71.5 & 78.7 & 62.5 & 27.8 & 4.0  & 7.8 \\
\midrule
\multirow{2}{*}{DeepSeek-V3.2 Chat}
  & EN & 84.6 & 20.9 & 28.2 & 48.9 & 68.0 & 40.3 & 15.3 & 6.7  & 7.8 \\
  & CN & 87.2 & 24.0 & 33.2 & 50.6 & 73.3 & 44.4 & 27.8 & 9.3  & 7.8 \\
\midrule
\multirow{2}{*}{DeepSeek-V3.2 Reasoner}
  & EN & 74.6 & 24.9 & 29.9 & 45.5 & 69.3 & 37.5 & 19.4 & 13.3 & 6.2 \\
  & CN & 76.8 & 22.9 & 29.3 & 45.5 & 69.3 & 40.3 & 18.1 & 10.7 & 4.7 \\
\bottomrule
\end{tabular}
\end{table}

%% file: tables/category.tex
\begin{table}[htbp]
\centering
\caption{Coverage of drug discovery workflow categories in MolViBench.}
\label{tab:workflow_coverage}
\begin{tabular}{@{}lcp{8cm}@{}}
\toprule
\rowcolor{headergray}
\textbf{Category} & \textbf{Tasks} & \textbf{Representative Operations} \\
\midrule
Molecular Characterization & $\sim$75 & MW, LogP, TPSA, fingerprints, scaffolds \\
ADMET Screening & $\sim$30 & BBB permeability, CYP450, hERG, Ames, PAINS \\
Virtual Screening & $\sim$25 & Tanimoto search, cascade filtering, scoring \\
Combinatorial Chemistry & $\sim$20 & R-group enumeration, BRICS recombination \\
Lead Optimization & $\sim$30 & LogP/TPSA tuning, QED optimization, SA score \\
Reaction Chemistry & $\sim$25 & Suzuki, Click, Buchwald--Hartwig, retrosynthesis \\
Clustering \& Selection & $\sim$15 & Butina, K-means, MaxMin, representative picking \\
QSAR / ML & $\sim$15 & Descriptor selection, RF regression, scaffold split \\
Molecular Generation & $\sim$15 & Derivative enumeration, genetic algorithm, quality metrics \\
Cheminformatics I/O & $\sim$20 & SDF/PDB parsing, CSV pipelines, batch processing \\
Stereochemistry & $\sim$15 & R/S assignment, E/Z detection, isomer enumeration \\
Rule-Based Filtering & $\sim$25 & Lipinski, Veber, Ghose, Egan, multi-rule cascades \\
\bottomrule
\end{tabular}

\end{table}

%% file: reference.bib
@article{wang2025cmphysbench,
  title={CMPhysBench: A Benchmark for Evaluating Large Language Models in Condensed Matter Physics},
  author={Wang, Weida and Huang, Dongchen and Li, Jiatong and Yang, Tengchao and Zheng, Ziyang and Zhang, Di and Han, Dong and Chen, Benteng and Luo, Binzhao and Liu, Zhiyu and others},
  journal={arXiv preprint arXiv:2508.18124},
  year={2025}
}

@article{guo2023can,
  title={What can large language models do in chemistry? a comprehensive benchmark on eight tasks},
  author={Guo, Taicheng and Nan, Bozhao and Liang, Zhenwen and Guo, Zhichun and Chawla, Nitesh and Wiest, Olaf and Zhang, Xiangliang and others},
  journal={Advances in neural information processing systems},
  volume={36},
  pages={59662--59688},
  year={2023}
}

@article{li2024empowering,
  title={Empowering molecule discovery for molecule-caption translation with large language models: A chatgpt perspective},
  author={Li, Jiatong and Liu, Yunqing and Fan, Wenqi and Wei, Xiao-Yong and Liu, Hui and Tang, Jiliang and Li, Qing},
  journal={IEEE Transactions on Knowledge and Data Engineering},
  year={2024},
  publisher={IEEE}
}

@inproceedings{solovev2025chemcoscientist,
  title={ChemCoScientist: LLM-Based Multi-Agent Assistant for Automated Solving of Chemical Tasks Using Data-Driven Tools},
  author={Solovev, Gleb V and Gurev, Ivan and Vepreva, Anastasia and Dubrovsky, Ivan and Zhidkovskaya, Alina and Fatkhiev, Kamil and Lutsenko, Elizaveta and Orlova, Anastasia and Gubina, Nina and Nikitin, Nikolay O and others},
  booktitle={2025 IEEE International Conference on Data Mining Workshops (ICDMW)},
  pages={2569--2572},
  year={2025},
  organization={IEEE}
}

@article{li2024speak,
  title={Speak-to-Structure: Evaluating LLMs in Open-domain Natural Language-Driven Molecule Generation},
  author={Li, Jiatong and Li, Junxian and Wang, Weida and Liu, Yunqing and Zheng, Changmeng and Zhou, Dongzhan and Wei, Xiao-yong and Li, Qing},
  journal={arXiv preprint arXiv:2412.14642},
  year={2024}
}

@article{m2024augmenting,
  title={Augmenting large language models with chemistry tools},
  author={M. Bran, Andres and Cox, Sam and Schilter, Oliver and Baldassari, Carlo and White, Andrew D and Schwaller, Philippe},
  journal={Nature machine intelligence},
  volume={6},
  number={5},
  pages={525--535},
  year={2024},
  publisher={Nature Publishing Group UK London}
}

@inproceedings{tang2025chemagent,
  title={Chemagent: Self-updating memories in large language models improves chemical reasoning},
  author={Tang, Xiangru and Hu, Tianyu and Ye, Muyang and Shao, Yanjun and Yin, Xunjian and Ouyang, Siru and Zhou, Wangchunshu and Lu, Pan and Zhang, Zhuosheng and Zhao, Yilun and others},
  booktitle={The Thirteenth International Conference on Learning Representations},
  year={2025}
}

@article{ivanovic2020lipinski,
  title={Lipinski’s rule of five, famous extensions and famous exceptions},
  author={Ivanovi{\'c}, Violeta and Ran{\v{c}}i{\'c}, Miroslav and Arsi{\'c}, Biljana and Pavlovi{\'c}, Aleksandra},
  journal={Popular Scientific Article},
  volume={3},
  number={1},
  pages={171--177},
  year={2020}
}

@article{randic1991generalized,
  title={Generalized molecular descriptors},
  author={Randi{\'c}, Milan},
  journal={Journal of Mathematical Chemistry},
  volume={7},
  number={1},
  pages={155--168},
  year={1991},
  publisher={Springer}
}

@article{noor2024deep,
  title={Deep learning pipeline for accelerating virtual screening in drug discovery},
  author={Noor, Fatima and Junaid, Muhammad and Almalki, Atiah H and Almaghrabi, Mohammed and Ghazanfar, Shakira and Tahir ul Qamar, Muhammad},
  journal={Scientific Reports},
  volume={14},
  number={1},
  pages={28321},
  year={2024},
  publisher={Nature Publishing Group UK London}
}

@article{krathwohl2002revision,
  title={A revision of Bloom's taxonomy: An overview},
  author={Krathwohl, David R},
  journal={Theory into practice},
  volume={41},
  number={4},
  pages={212--218},
  year={2002},
  publisher={Taylor \& Francis}
}

@article{chen2021evaluating,
  title={Evaluating large language models trained on code},
  author={Chen, Mark and Tworek, Jerry and Jun, Heewoo and Yuan, Qiming and Pinto, Henrique Ponde De Oliveira and Kaplan, Jared and Edwards, Harri and Burda, Yuri and Joseph, Nicholas and Brockman, Greg and others},
  journal={arXiv preprint arXiv:2107.03374},
  year={2021}
}

@article{austin2021program,
  title={Program synthesis with large language models},
  author={Austin, Jacob and Odena, Augustus and Nye, Maxwell and Bosma, Maarten and Michalewski, Henryk and Dohan, David and Jiang, Ellen and Cai, Carrie and Terry, Michael and Le, Quoc and others},
  journal={arXiv preprint arXiv:2108.07732},
  year={2021}
}

@inproceedings{jimenezswe,
  title={SWE-bench: Can Language Models Resolve Real-world Github Issues?},
  author={Jimenez, Carlos E and Yang, John and Wettig, Alexander and Yao, Shunyu and Pei, Kexin and Press, Ofir and Narasimhan, Karthik R},
  booktitle={The Twelfth International Conference on Learning Representations},
  year={2024}
}

@inproceedings{haobeyond,
  title={Beyond Chemical QA: Evaluating LLM's Chemical Reasoning with Modular Chemical Operations},
  author={Hao, Li and He, CAO and Feng, Bin and Shao, Daniel and Tang, Xiangru and Yan, Zhiyuan and Tian, Yonghong and Yuan, Li and Li, Yu},
  booktitle={The Thirty-ninth Annual Conference on Neural Information Processing Systems Datasets and Benchmarks Track},
  year={2025}
}

@article{landrum2013rdkit,
  title={Rdkit documentation},
  author={Landrum, Greg and others},
  journal={Release},
  volume={1},
  number={1-79},
  pages={4},
  year={2013}
}

@inproceedings{jainlivecodebench,
  title={LiveCodeBench: Holistic and Contamination Free Evaluation of Large Language Models for Code},
  author={Jain, Naman and Han, King and Gu, Alex and Li, Wen-Ding and Yan, Fanjia and Zhang, Tianjun and Wang, Sida and Solar-Lezama, Armando and Sen, Koushik and Stoica, Ion},
  booktitle={The Thirteenth International Conference on Learning Representations},
  year={2025}
}

@inproceedings{zhuobigcodebench,
  title={BigCodeBench: Benchmarking Code Generation with Diverse Function Calls and Complex Instructions},
  author={Zhuo, Terry Yue and Chien, Vu Minh and Chim, Jenny and Hu, Han and Yu, Wenhao and Widyasari, Ratnadira and Yusuf, Imam Nur Bani and Zhan, Haolan and He, Junda and Paul, Indraneil and others},
  booktitle={The Thirteenth International Conference on Learning Representations},
  year={2025}
}

@inproceedings{lai2023ds,
  title={DS-1000: A natural and reliable benchmark for data science code generation},
  author={Lai, Yuhang and Li, Chengxi and Wang, Yiming and Zhang, Tianyi and Zhong, Ruiqi and Zettlemoyer, Luke and Yih, Wen-tau and Fried, Daniel and Wang, Sida and Yu, Tao},
  booktitle={International Conference on Machine Learning},
  pages={18319--18345},
  year={2023},
  organization={PMLR}
}

@article{wu2018moleculenet,
  title={MoleculeNet: a benchmark for molecular machine learning},
  author={Wu, Zhenqin and Ramsundar, Bharath and Feinberg, Evan N and Gomes, Joseph and Geniesse, Caleb and Pappu, Aneesh S and Leswing, Karl and Pande, Vijay},
  journal={Chemical science},
  volume={9},
  number={2},
  pages={513--530},
  year={2018},
  publisher={Royal Society of Chemistry}
}

@inproceedings{fangmol,
  title={Mol-Instructions: A Large-Scale Biomolecular Instruction Dataset for Large Language Models},
  author={Fang, Yin and Liang, Xiaozhuan and Zhang, Ningyu and Liu, Kangwei and Huang, Rui and Chen, Zhuo and Fan, Xiaohui and Chen, Huajun},
  booktitle={The Twelfth International Conference on Learning Representations},
  year={2024}
}

@misc{anthropic_claude_2026,
  title        = {Introducing Claude Opus 4.6},
  author       = {Anthropic},
  year         = {2026},
  howpublished = {\url{https://www.anthropic.com/news/claude-opus-4-6}},
}

@article{liu2025deepseek,
  title={Deepseek-v3. 2: Pushing the frontier of open large language models},
  author={Liu, Aixin and Mei, Aoxue and Lin, Bangcai and Xue, Bing and Wang, Bingxuan and Xu, Bingzheng and Wu, Bochao and Zhang, Bowei and Lin, Chaofan and Dong, Chen and others},
  journal={arXiv preprint arXiv:2512.02556},
  year={2025}
}

@misc{openai_codex52,
  title        = {Introducing GPT‑5.2‑Codex},
  author       = {OpenAI},
  year         = {2025},
  howpublished = {\url{https://openai.com/index/introducing-gpt-5-2-codex}},
}

@misc{openai_codex53,
  title        = {Introducing GPT‑5.3‑Codex},
  author       = {OpenAI},
  year         = {2026},
  howpublished = {\url{https://openai.com/index/introducing-gpt-5-3-codex}},
}

@article{team2026kimi,
  title={Kimi K2. 5: Visual Agentic Intelligence},
  author={Team, Kimi and Bai, Tongtong and Bai, Yifan and Bao, Yiping and Cai, SH and Cao, Yuan and Charles, Y and Che, HS and Chen, Cheng and Chen, Guanduo and others},
  journal={arXiv preprint arXiv:2602.02276},
  year={2026}
}

@misc{minimax25,
  title        = {MiniMax M2.5: Built for Real-World Productivity.},
  author       = {MiniMax},
  year         = {2026},
  howpublished = {\url{https://www.minimax.io/news/minimax-m25}},
}

@misc{gemini3,
  title        = {A new era of intelligence with Gemini 3.},
  author       = {Google DeepMind},
  year         = {2025},
  howpublished = {\url{https://blog.google/products-and-platforms/products/gemini/gemini-3/#gemini-3}},
}

@article{sterling2015zinc,
  title={ZINC 15--ligand discovery for everyone},
  author={Sterling, Teague and Irwin, John J},
  journal={Journal of chemical information and modeling},
  volume={55},
  number={11},
  pages={2324--2337},
  year={2015},
  publisher={ACS Publications}
}

@article{tang2023biocoder,
  title={BioCoder: A Benchmark for Bioinformatics Code Generation with Large Language Models},
  author={Tang, Xiangru and Qian, Bill and Gao, Rick and Chen, Jiakang and Chen, Xinyun and Gerstein, Mark},
  journal={arXiv preprint arXiv:2308.16458},
  year={2023}
}

@inproceedings{chenscienceagentbench,
  title={ScienceAgentBench: Toward Rigorous Assessment of Language Agents for Data-Driven Scientific Discovery},
  author={Chen, Ziru and Chen, Shijie and Ning, Yuting and Zhang, Qianheng and Wang, Boshi and Yu, Botao and Li, Yifei and Liao, Zeyi and Wei, Chen and Lu, Zitong and others},
  booktitle={The Thirteenth International Conference on Learning Representations},
  year={2025}
}
